\documentclass[10pt,twocolumn,letterpaper]{article}
\usepackage{cvpr}

\usepackage{booktabs}
\usepackage{float}
\usepackage{makecell}
\usepackage{multirow}
\usepackage{pifont}
\usepackage{textpos}
\usepackage{xspace}
\usepackage{colortbl} 
\usepackage{graphicx}   %
\usepackage{caption}    %
\usepackage{subcaption}

\definecolor{cvprblue}{rgb}{0.21,0.49,0.74}
\usepackage[pagebackref,breaklinks,colorlinks,citecolor=cvprblue]{hyperref}

\newcolumntype{x}[1]{>{\centering\arraybackslash}p{#1pt}}
\newlength\savewidth
\newcommand{\tablestyle}[2]{\setlength{\tabcolsep}{#1}\renewcommand{\arraystretch}{#2}\centering\footnotesize}

\newcommand{\mname}{Tracktention\xspace}

\makeatletter
\renewcommand{\paragraph}{%
  \@startsection{paragraph}{4}%
  {\z@}{-0.2em}{-0.5em}%
  {\normalfont\normalsize\bfseries}%
}
\makeatother

\def\paperID{12604}
\def\confName{CVPR}
\def\confYear{2025}

\title{\mname: Leveraging Point Tracking to Attend Videos Faster and Better}

\author{Zihang Lai \hspace{2em} Andrea Vedaldi\\
Visual Geometry Group (VGG), University of Oxford\\
{\tt\small \{zlai,vedaldi\}@robots.ox.ac.uk}
}

\begin{document}
\twocolumn[{
\maketitle
\centering
\begin{minipage}[t]{0.65\textwidth}\vspace{0pt}
\includegraphics[width=\textwidth]{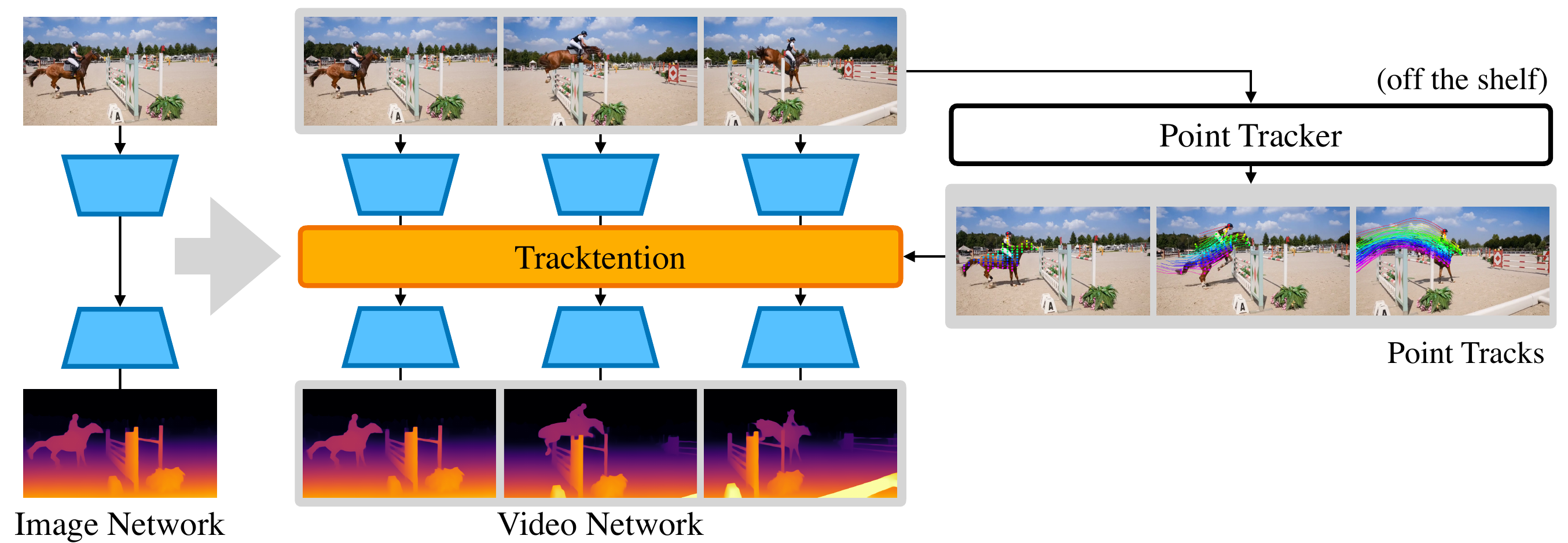}
\end{minipage}\hfill%
\begin{minipage}[t]{0.34\textwidth}\vspace{0pt}
\includegraphics[width=\textwidth]{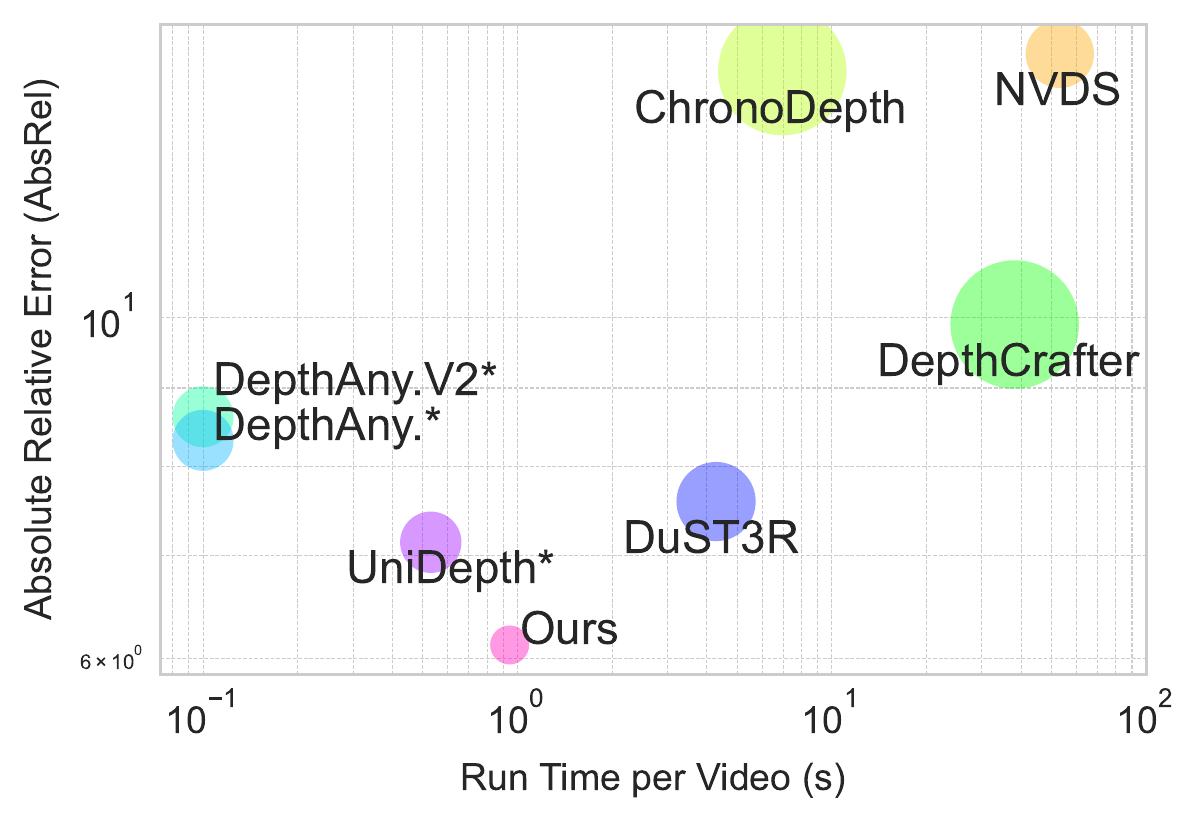}
\end{minipage}\vspace{-1em}
\captionsetup{type=figure}
\captionof{figure}{
\emph{Left}:
The \textbf{\mname Layer} is a plug-and-play module that can convert an image-based network (\eg, for monocular depth prediction) into a state-of-the-art video network (\eg, for video depth prediction).
It does so by integrating the output of any off-the-shelf, modern, and powerful point trackers via track cross-attention.
\emph{Right}:
For example, \mname achieves \textbf{state-of-the-art and efficient video depth} prediction by transforming Depth Anything into a video depth model.
See \cref{tab:depth_robustmvd} for detailed results. ${}^\ast$Single-image models.
}%
\label{fig:teaser}
\vspace{3em}
}]

\begin{abstract}
Temporal consistency is critical in video prediction to ensure that outputs are coherent and free of artifacts.
Traditional methods, such as temporal attention and 3D convolution, may struggle with significant object motion and may not capture long-range temporal dependencies in dynamic scenes.
To address this gap, we propose the \mname Layer, a novel architectural component that explicitly integrates motion information using point tracks, \ie, sequences of corresponding points across frames.
By incorporating these motion cues, the \mname Layer enhances temporal alignment and effectively handles complex object motions, maintaining consistent feature representations over time.
Our approach is computationally efficient and can be seamlessly integrated into existing models, such as Vision Transformers, with minimal modification.
It can be used to upgrade image-only models to state-of-the-art video ones, sometimes outperforming models natively designed for video prediction.
We demonstrate this on video depth prediction and video colorization, where models augmented with the \mname Layer exhibit significantly improved temporal consistency compared to baselines.
Project website: \url{zlai0.github.io/TrackTention}.
\end{abstract}

\section{Introduction}%
\label{sec:intro}

\begin{figure*}[ht]
\centering
\includegraphics[width=0.92\textwidth]{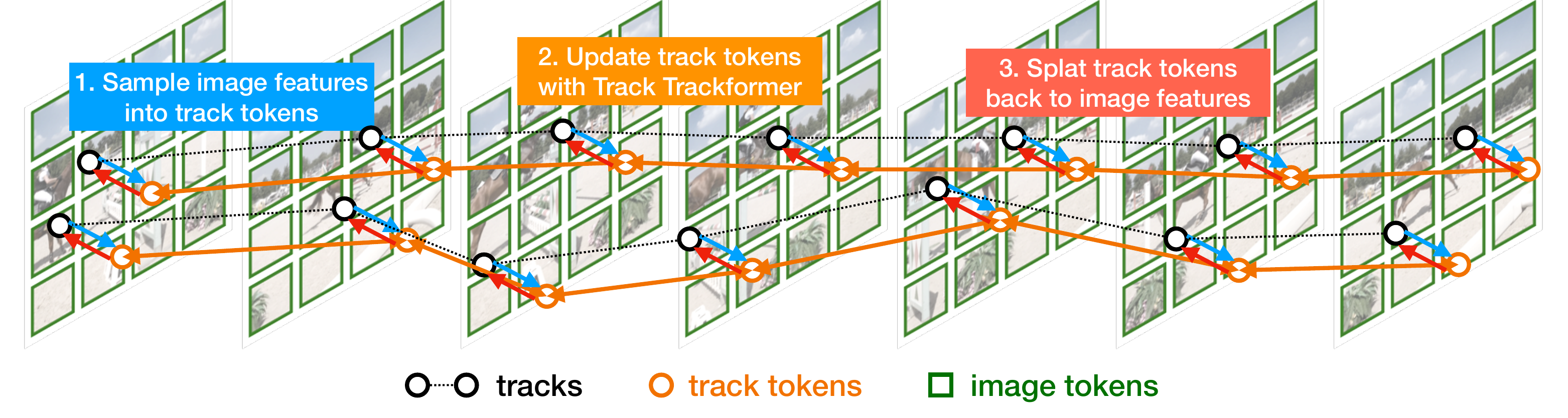}
\vspace{-1em}
\caption{
\textbf{Overview of \mname}.
We begin by using an off-the-shelf point tracker to extract a number of video tracks.
Given these, we first 
\textcolor[rgb]{0.282, 0.624, 0.973}{\textbf{sample}} image tokens at the track locations, obtaining corresponding track tokens (\cref{sec:attentional-sampling}).
Next, we use a Track Transformer to \textcolor[rgb]{0.941, 0.596, 0.216}{\textbf{update}}
these tokens, propagating information temporally at consistent spatial locations (\cref{sec:track-transformer}).
Finally, we \textcolor[rgb]{0.855, 0.231, 0.149}{\textbf{splat}} the information back to the image tokens (\cref{sec:attentional-splatting}).
By explicitly incorporating motion information through point tracks, \mname improves temporal alignment, effectively captures complex object movements, and ensures stable feature representations over time.
}%
\label{fig:teaser_p3}
\vspace{-1em}
\end{figure*}

Compared to image analysis tasks, video analysis tasks, such as video segmentation, video depth estimation, and video colorization, pose additional challenges due to the temporal dimension inherent in video data.
A particularly important challenge is ensuring that outputs are temporally consistent across frames, which is necessary to produce coherent and artifact-free results.
Temporal inconsistencies, such as flickering or abrupt changes in predicted object attributes, are especially noticeable in applications like colorization, where the output is a new video.

An aspect of video analysis that has seen significant progress in recent years is \emph{point tracking}.
New trackers such as PIPs~\cite{harley22particle}, TAPIR~\cite{doersch23tapir:}, BootsTAP~\cite{doersch24bootstap:}, and CoTracker~\cite{karaev24cotracker} can now track quasi-dense collections of points across long video sequences with high reliability and efficiency.
So far, these trackers have been primarily used in selected applications like 3D reconstruction and object tracking through ad-hoc algorithms.
In this paper, we ask whether progress in point tracking can benefit a much wider range of video analysis tasks.

To answer this question, we propose the \emph{\mname Layer}, a novel architectural component designed to enhance temporal consistency in video transformers.
\mname integrates seamlessly into vision transformers as a new attention layer, making it compatible with a wide range of existing models.
The key innovation of \mname lies in leveraging pre-extracted point tracks to inform the model about temporal correspondences between image tokens in different frames (\cref{fig:teaser_p3}).
This is achieved by sampling the image tokens at the track points.
The information is then propagated \emph{along} the tracks using a transformer layer, which justifies the name ``\mname''.
Finally, the information is \emph{splatted} back to the image tokens, allowing the computation to progress as normal.

By explicitly incorporating tracking information, the model can establish direct correspondences, enabling precise temporal alignment regardless of how much objects move.
This explicit temporal alignment allows the model to maintain consistent feature representations over time, smoothing features, and reducing temporal artifacts.

The problem of propagating information in video neural networks has been extensively studied in the literature.
Many approaches are \emph{implicit}, in the sense that they do not explicitly estimate or account for the motion observed in the scene.
Examples include 3D convolution~\cite{tran2015learning,carreira2017quo} and space-time attention~\cite{wang18non-local,bertasius2021space,arnab2021vivit,neimark21video,fan21multiscale}.
However, to maintain manageable computational costs, these methods often reduce the spatial resolution of features and limit the temporal range they can model---either by using small convolution kernels or by decoupling spatial and temporal attention.
The resulting networks may struggle to represent motion precisely and to keep up with large object displacements.

In contrast, \mname explicitly accounts for motion at the resolution of point tracks, which is much finer than that of typical image features.
By pooling and splatting information at image locations that are \emph{already} put in correspondence by the tracker, it only needs to propagate information along the tracks, obviating the need to implicitly establish such correspondences via space-time attention.

There are also approaches to video analysis that, like \mname, incorporate motion information \emph{explicitly}.
They often work by feeding optical flow to the network, capturing pixel-level motion between frames~\cite{patrick21keeping}.
While optical flow can be effective in certain contexts, it may struggle with occlusions or large displacements.
Building on the power of point trackers, \mname can handle occlusions, large displacements, and long temporal dependencies more effectively.

In summary, \mname offers several advantages over traditional video network architectures:
(1) It leverages the power of modern point tracker models, which are excellent \emph{motion experts}, and provides a general mechanism to inject this knowledge into any video transformer.
In particular, it can handle complex and large object motions, limited only by the capabilities of the point trackers.
(2) It directly establishes long-term space-time correspondences, unlike 3D convolutions that are local and unlike separated space-time attention, which can only do so indirectly.
(3) It avoids the need to repeatedly calculate such correspondences explicitly, as full spatio-temporal attention does, achieving much greater efficiency and operating at a much finer resolution.

We deliver \mname as a \textbf{plugin layer} (\cref{fig:teaser}) that can be added to single-image neural network architectures to extend them to powerful, temporally-aware video models.
For example, we demonstrate that a pre-trained single-image depth predictor can be extended to a corresponding \emph{state-of-the-art} \emph{video depth predictor} by integrating our \mname Layer.
We achieve similar results for \emph{automatic video colorization}, outperforming models natively designed to process videos while also being more efficient.

\section{Related Work}%
\label{sec:related}

\paragraph{Temporal Modeling for Videos.}

Temporal modeling is a cornerstone of video understanding tasks.
Over the years, various techniques have been proposed to capture temporal dependencies in video data.
3D CNNs~\cite{tran2015learning, carreira2017quo, feichtenhofer2019slowfast} extend 2D convolutions into the temporal dimension, allowing for direct spatiotemporal feature extraction from video data.
Temporal attention mechanisms provide a new approach to effectively capture long-range dependencies across frames.
Video transformers~\cite{bertasius2021space, arnab2021vivit, fan2021multiscale, liu2022video, patrick21keeping} apply self-attention mechanisms over temporal and spatial dimensions, typically separately or hierarchically, to reduce computational complexity while modeling temporal dependencies effectively.
Recent video processing models~\cite{blattmann2023align, singer2022make, blattmann2023stable} also use a hybrid of these approaches.
However, these methods have notable limitations: temporal attention struggles with rapid object movements, while 3D convolutions assume local spatiotemporal correlations, failing with unpredictable motion.
In this work, we explicitly leverage tracking information to precisely align temporal features regardless of object movement.

\paragraph{Consistent Video Prediction.}

Addressing temporal inconsistency is critical for ensuring smooth and coherent video outputs.
Optical flow-based methods~\cite{zhu2017deep} enforce temporal coherence by aligning features or predictions across frames, propagating features along motion paths estimated by optical flow.
Liu et al.~\cite{liu2017video} utilize convolutional LSTMs to capture temporal information for consistent video frame synthesis, modeling temporal dependencies through recurrent connections.
Lai et al.~\cite{lai2018learning} propose a post-processing model with a temporal consistency objective for video prediction.
Similarly, Luo et al.~\cite{luo2020consistent} and Kopf et al.~\cite{kopf2021robust} present test-time training post-processing methods specifically designed to stabilize video depth predictions using geometric consistency cues.
In contrast, our work aims to \emph{stabilize features} by allowing them to attend to corresponding areas across time based on point tracks.
This approach inherently accounts for object movement and embeds temporal alignment efficiently within the model.

\paragraph{Point Tracking.}

Point tracking is the task of identifying and continuously tracking specific points across frames in a video.
PIPs~\cite{harley2022particle} pioneered a neural network-based approach that updates tracked points iteratively by extracting correlation maps between frames, inspired by the RAFT model for optical flow~\cite{teed2020raft}.
TAP-Vid~\cite{doersch2022tap} proposed a comprehensive benchmark and TAP-Net, a new model for point tracking.
TAPIR~\cite{doersch2023tapir} further improved performance by combining global matching capabilities with PIPs, enhancing the accuracy of tracked points even in complex motion scenarios.
OmniMotion~\cite{wang2023tracking} tracks points by leveraging a quasi-3D canonical volume but is approximately $10^3$ times slower than CoTracker due to its test-time optimization.
CoTracker~\cite{karaev23cotracker} leveraged a transformer-based architecture to track multiple points jointly, yielding improvements particularly for occluded points.
Recent methods such as LocoTrack~\cite{cho2024local}, which introduced 4D correlation features, and TAPTR~\cite{li2025taptr} further improved precision.

In this work, we do not propose new models for point tracking.
Instead, we use existing trackers to establish direct token correspondences, ensuring precise temporal alignment despite object motion.

\section{Method}%
\label{sec:method}

\begin{figure*}[ht]
\centering
\includegraphics[width=0.99\textwidth]{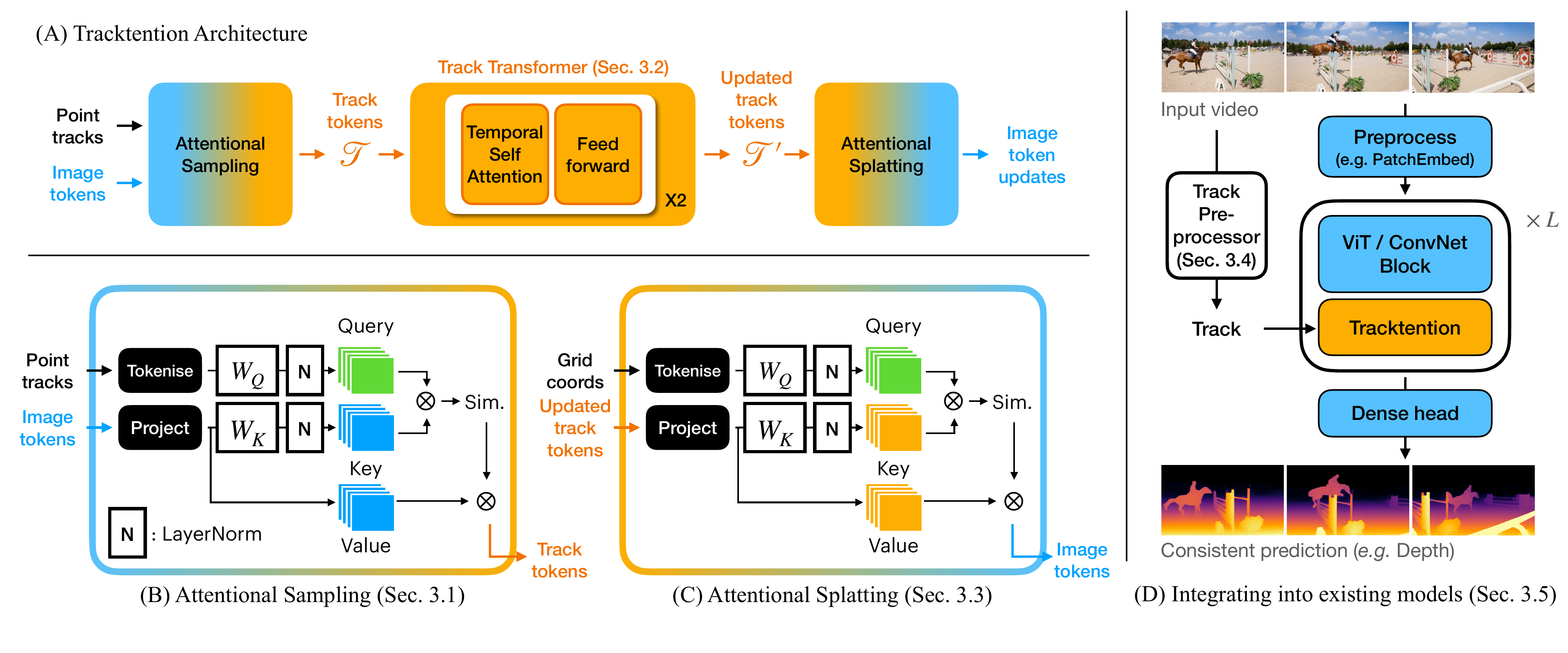}
\vspace{-2em}
\caption{
Left: the \textbf{\mname architecture} comprises Attentional Sampling, pooling information from images to track, Track Transformer, processing this information temporally, and Attentional Splatting, moving the processed information back to the images. 
Right: \mname is easily \textbf{integrated} in ViTs and ConvNets to make video networks out image ones.}%
\label{fig:mainfig}
\vspace{-1em}
\end{figure*}

In this section, we introduce \mname, a novel transformer layer designed to aggregate and redistribute feature information in video transformers in a motion-aware manner (\cref{fig:mainfig}).
By leveraging motion information captured by the output of a point tracker, this layer improves the temporal consistency and motion awareness of video features.

\mname comprises an Attentional Sampling block (\cref{sec:attentional-sampling}) that pools information from the video features into the tracks, a Track Transformer (\cref{sec:track-transformer}) that updates the track tokens, and an Attentional Splatting block (\cref{sec:attentional-splatting}) that pushes the updated track tokens back to the video features.
We describe these components next.

\subsection{Attentional Sampling}%
\label{sec:attentional-sampling}

\newcommand{\Ntracks}{M}

The \emph{Attentional Sampling} module is the first stage in \mname.
It uses cross-attention to pool information from the video features into the track tokens.

Let $F \in \mathbb{R}^{T \times HW \times D_f}$ denote the input feature map, where $T$ is the number of time steps, $H$ and $W$ are the spatial dimensions, and $D_f$ is the feature dimension.
Let $\mathbf{P} \in \mathbb{R}^{T \times \Ntracks \times 2}$ represent a set of $\Ntracks$ tracks, each of which is a sequence of 2D points.
We first apply a positional embedding to these 2D points, obtaining one token per track and per frame, and arrange these tokens into a tensor $\mathcal{T} \in \mathbb{R}^{T \times \Ntracks \times D_f}$.
We then project the track tokens into queries $Q \in \mathbb{R}^{T \times \Ntracks \times D_k}$ and the feature map into keys $K \in \mathbb{R}^{T \times HW \times D_k}$ and values $V \in \mathbb{R}^{T \times HW \times D_f}$ as:
\[
Q_t = \mathcal{T}_t W_Q, \quad K_t = F_t W_K, \quad V_t = F_t,
\]
where $W_Q$ and $W_K$ are learnable weight matrices, and the subscript $t \in \{1, \dots, T\}$ indexes the first dimensions of the tensors, extracting matrices.
Note that we avoid projecting the values to preserve the original features during sampling.

For each time $t$, the attention weights $A_t \in \mathbb{R}^{\Ntracks \times HW}$ are computed using the scaled dot-product attention mechanism:
\begin{equation}
\label{eq:attention}
A_t
=
\operatorname{softmax}
\left(
    \frac{Q_t K_t^\top}{\sqrt{D_k}} + B_t
\right),
\end{equation}
where $B_t \in \mathbb{R}^{\Ntracks \times HW}$ is a bias term that encourages attention to align with the tracks.
It is defined as:
\[
B_{tij} = -
\frac{\|P_{ti} - \operatorname{pos}_{F_t}(j)\|^2}{2 \sigma^2},
\]
where $i \in \{ 1, \dots, \Ntracks \}$, $j \in \{1, \dots, HW\}$, $P_{ti} \in \mathbb{R}^2$ is the spatial location of track $i$ at time $t$, and $\operatorname{pos}_{F_t}(j) \in \mathbb{R}^2$ is the spatial location of the $j$-th feature map token at time $t$.

In the softmax operator of \cref{eq:attention}, this bias is analogous to multiplying the attention map by a Gaussian window that decays exponentially as the feature position deviates from the track position.
The parameter $\sigma$, which we set to $1/2$, determines how fast the window decays.

We use the attention to ``sample'' track features $S \in \mathbb{R}^{T \times \Ntracks \times D_f}$ by setting $S_t = A_t V_t = A_t F_t \in \mathbb{R}^{\Ntracks \times D_f}$ for each time $t$.
Alternatively, we could have obtained $S$ by \emph{sampling} the features $F$ at the given track locations, for instance, by using bilinear sampling.
This can also be expressed as $A_t F_t$ for a specific matrix $A_t$, so sampling is a special case of our Attentional Sampling block.
The advantage of Attentional Sampling is that it allows the model to learn a better sampling strategy.

\paragraph{Relative Spatial Information.}%
\label{sec:relative-spatial-information}

To capture the relative spatial relationships between track tokens and feature map locations, we incorporate Rotational Position Encodings (RoPE)~\cite{su2021roformer} into the keys and values derived from the feature map.
RoPE is a form of positional encoding that allows attention mechanisms to be aware of the relative positions of points defined in a continuous space.

To compute the RoPE embedding $\text{RoPE}(\mathbf{f})$ of a feature vector $\mathbf{f}$ at spatial location $(x, y)$, we use the first $d/2$ channels to encode the $x$ position.
For $i = 1, \dots, d/4$, we define:
\begin{align*}
\text{RoPE}(\mathbf{f})[2i] &= \mathbf{f}[2i] \cdot \cos(\theta_i x) - \mathbf{f}[2i+1] \cdot \sin(\theta_i x), \\
\text{RoPE}(\mathbf{f})[2i+1] &= \mathbf{f}[2i] \cdot \sin(\theta_i x) + \mathbf{f}[2i+1] \cdot \cos(\theta_i x),
\end{align*}
where $\theta_i = 100^{\frac{-2(i-1)}{d/2}}$.
The second $d/2$ channels encode the $y$ position analogously by replacing $x$ with $y$ in the above equations.

\paragraph{Implementation Details.}

To stabilize the training process, especially due to the difference in scale between the track token embeddings and the feature map embeddings (which tend to have larger norms), we apply QK-normalization~\cite{dehghani2023scaling} to the queries and keys before computing the attention weights.
This normalization ensures that the scale of the embeddings is consistent, preventing issues related to gradient instability.
We also use $H$ attention heads, allowing the model to capture diverse contextual relationships across different subspaces.

\subsection{Track Transformer}%
\label{sec:track-transformer}

Once the track tokens $S$ are obtained, the \emph{Track Transformer} processes them to propagate information along the tracks, making them more temporally consistent and smoothing out any irregularities.
It is designed as a transformer that operates along the time dimension of the point tracks and outputs updated track tokens $\mathcal{T}'$.

In more detail, recall that the track tokens $S$ are of shape $T \times \Ntracks \times D_f$.
We swap the first two dimensions to obtain a tensor of shape $\Ntracks \times T \times D_f$.
Then, we apply a transformer block, interpreting the first dimension $\Ntracks$ as the batch dimension.
The effect is to apply attention along the temporal dimension, allowing the model to learn temporal dependencies along each track.

This approach ensures that information is not exchanged across tracks, meaning each track's temporal sequence is processed independently.
We considered a variant of the Track Transformer that included attention across tracks as well, but found that this approach was slightly worse and more computationally expensive.
Our interpretation is that information is already exchanged spatially by the Vision Transformer paired with \mname, making additional exchange of information across locations redundant.

\paragraph{Architecture.}

For each track, we utilize a 2-layer transformer encoder along the temporal dimension, structured as a standard transformer model.
Each layer consists of Multi-Head Self-Attention, which allows each time step within a track to attend to all other time steps within that same track, and Position-Wise Feed-Forward Networks, which apply nonlinear transformations independently to each time step, enhancing the model's representational capacity.
We also incorporate sinusoidal positional encodings to provide the model with a sense of temporal order.
These encodings are added to the input features prior to the transformer layers, ensuring the model can distinguish between different time steps and understand the temporal structure of the sequence.

\subsection{Attentional Splatting}%
\label{sec:attentional-splatting}

The \emph{Attentional Splatting} module maps the updated track tokens $\mathcal{T}'$ back onto the feature maps $F$.
This is achieved by reversing the roles of the track tokens and the feature tokens in the various expressions in \cref{sec:attentional-sampling}.
This means the queries are derived from the grid coordinates of the output feature map, while the keys and values are generated from the track tokens.
The same bias term $B_t$ is used as before (but now the matrix is transposed), as well as the same $QK$-normalization and RoPE encodings, this time obtaining a matrix $A_t' \in \mathbb{R}^{HW \times \Ntracks}$.
The final output $\operatorname{\mname}(F)$ is then computed as:
\[
[\operatorname{\mname}(F)]_t
=
W_{\text{out}} A_t'[\mathcal{T}']_t,
\]
where $W_{\text{out}}$ is a final output projection.
By adopting a symmetric design for Attentional Sampling and Splatting, we ensure that information is handled in a consistent manner.

\subsection{Point Tracker Pre-Processor}%
\label{sec:pre-processing}

To apply the \mname{} layer to a video, we first need to extract the point tracks $\mathbf{P}$.
We use CoTracker3~\cite{karaev2024cotracker3} due to its robustness in handling complex motions and occlusions in videos.
Trackers require a seed point for each track, called a query.
We sample 576 points uniformly at random from the spatio-temporal video volume and track each forward and backward in time.
Compared to initializing tracks on a grid in the first frame of the video, this simple initialization scheme encourages tracks to cover the video well, despite camera and object motion.

\subsection{Integration into Image-Based Models}
A key application of \mname is to build video neural networks from image-based ones.
Integrating \mname into existing image-based models, such as a ViT or ConvNet, is straightforward.
For example, monocular depth estimation models like Depth Anything, which are designed for single images, may lack temporal consistency when applied to video frame by frame.
Adding \mname layers after each transformer block and fine-tuning the model on videos can increase temporal stability and significantly improve performance in video-based applications.

To incorporate \mname into an existing network backbone, we insert it immediately after all or some transformer or convolutional blocks.
We also include a residual connection to preserve the original information flow, setting
$
F' = F + \operatorname{\mname}(F),
$
where $F$ represents the output feature map from the preceding block, and $F'$ is the updated feature map after the \mname module.

Furthermore, the \mname output projection, $W_{\text{out}}$, is initialized to zero, preserving the original network output at training onset. 
This strategy retains benefits of pre-trained weights while allowing the model to gradually adapt to temporal updates. 
Consequently, \mname modules capture fine spatial details and ensure consistent temporal modeling across layers.

\section{Experiments}%
\label{sec:experiments}

\begin{table*}[t]
\small
\centering
\tablestyle{4.5pt}{0.95}
\begin{tabular}{lcc cc cc cc cc cc}
\toprule
\multirow{2}{*}{Method} & Type & \multirow{2}{*}{\#Params} & \multicolumn{2}{c}{Sintel ($\sim$50 frames)} & \multicolumn{2}{c}{Scannet (90 frames)} & \multicolumn{2}{c}{KITTI (110 frames)} & \multicolumn{2}{c}{Bonn (110 frames)} & \multicolumn{2}{c}{Average} \\
\cmidrule(lr){4-5}
\cmidrule(lr){6-7}
\cmidrule(lr){8-9}
\cmidrule(lr){10-11}
\cmidrule(lr){12-13}
& & & AbsRel $\downarrow$ & $\delta_{1.25} \uparrow$ & AbsRel $\downarrow$ & $\delta_{1.25} \uparrow$ & AbsRel $\downarrow$ & $\delta_{1.25} \uparrow$ & AbsRel $\downarrow$ & $\delta_{1.25} \uparrow$ & AbsRel $\downarrow$ & $\delta_{1.25} \uparrow$ \\
\midrule
DUSt3R~\cite{wang24dust3r:}            & Vid. & 578M  & 0.628             & 0.393                    & 0.194          & 0.694          & 0.292          & 0.456          & 0.250          & 0.588          & 0.341          & 0.533          \\
NVDS~\cite{wang2023neural}             & Vid. & 430M  & 0.408             & 0.483                    & 0.187          & 0.677          & 0.253          & 0.588          & 0.167          & 0.766          & 0.254          & 0.629          \\
ChronoDepth~\cite{shao24learning}      & Vid. & 1521M & 0.587             & 0.486                    & 0.159          & 0.783          & 0.167          & 0.759          & 0.100          & 0.911          & 0.253          & 0.735          \\
DepthCrafter~\cite{hu2024depthcrafter} & Vid. & 1521M & {0.343}$^\dagger$ & \textbf{0.673}$^\dagger$ & 0.125          & 0.848          & 0.110          & 0.881          & 0.075          & \textbf{0.971} & 0.163          & 0.843          \\
\midrule
Marigold~\cite{ke2024repurposing}      & Im.  & 865M  & 0.532             & 0.515                    & 0.166          & 0.769          & 0.149          & 0.796          & 0.091          & 0.931          & 0.235          & 0.753          \\
DepthAny.~\cite{yang24depth}~($\ast$)  & Im.  & 343M  & 0.325             & 0.564                    & 0.130          & 0.838          & 0.142          & 0.803          & 0.078          & 0.939          & 0.169          & 0.786          \\
DepthAny.-V2~\cite{yang24depthv2}      & Im.  & 343M  & 0.367             & 0.554                    & 0.135          & 0.822          & 0.140          & 0.804          & 0.106          & 0.921          & 0.187          & 0.775          \\
\midrule
\textbf{Ours}                          & Vid. & 140M  & \textbf{0.295}    & 0.640                    & \textbf{0.087} & \textbf{0.933} & \textbf{0.104} & \textbf{0.903} & \textbf{0.066} & \textbf{0.971} & \textbf{0.138} & \textbf{0.862} \\
\bottomrule
\end{tabular}
\vspace{-1em}
\caption{
\textbf{DepthCrafter~\cite{hu2024depthcrafter} video depth benchmark.}
Our model upgrades DepthAnything ($\ast$) to a video depth predictor, outperforming all baselines (image- or video-based) while having the smallest parameter count (140M).
We use DepthAnything-Base (97M) as the base model and still outperform the DepthAnything-Large (343M) model.
$^\dagger$Reproduced results confirmed by original authors.}%
\label{tab:depth_depthcrafter}
\vspace{-1em}
\end{table*}

\begin{table*}[ht]
    \small
    \centering
    \tablestyle{3.1pt}{0.95}
\begin{tabular}{l c c c c c c c c c c c c c c c c}
\toprule
\multirow{2}{*}{Methods} & \multirow{2}{*}{\makecell{GT \\ Pose}} & 
\multirow{2}{*}{\makecell{GT \\ Intr.}}  & \multirow{2}{*}{Align} & \multicolumn{2}{c}{KITTI} & \multicolumn{2}{c}{ScanNet} & \multicolumn{2}{c}{ETH3D} & \multicolumn{2}{c}{DTU} & \multicolumn{2}{c}{T\&T} & \multicolumn{2}{c}{Average} & \multirow{2}{*}{\makecell{Runtime \\ (s)}} \\
\cmidrule(lr){5-6} \cmidrule(lr){7-8} \cmidrule(lr){9-10} \cmidrule(lr){11-12} \cmidrule(lr){13-14} \cmidrule(lr){15-16}
& & & & rel $\downarrow$ & $\delta_{1.03}$ $\uparrow$ & rel $\downarrow$ & $\delta_{1.03}$ $\uparrow$ & rel $\downarrow$ & $\delta_{1.03}$ $\uparrow$ & rel $\downarrow$ & $\delta_{1.03}$ $\uparrow$ & rel $\downarrow$ & $\delta_{1.03}$ $\uparrow$ & rel $\downarrow$ & $\delta_{1.03}$ $\uparrow$ & \\
\midrule
COLMAP~\cite{schonberger16structure-from-motion, schonberger16pixelwise}$^\sharp$ & \ding{51} & \ding{51} & $\times$ & \textbf{12.0} & \textbf{58.2} & \textbf{14.6} & \textbf{34.2} & \textbf{16.4} & \textbf{55.1} & \textbf{0.7} & \textbf{96.5} & \textbf{2.7} & \textbf{95.0} & \textbf{9.3} & \textbf{67.8}
 & $\approx$ 3 min\\
 COLMAP Dense~\cite{schonberger16structure-from-motion, schonberger16pixelwise} & \ding{51} & \ding{51} & $\times$ & 26.9 & 52.7 & 38.0 & 22.5 & 89.8 & 23.2 & 20.8 & 69.3 & 25.7 & 76.4 & 40.2 & 48.8 & $\approx$ 3 min\\
\midrule
NVDS~\cite{wang23neural} & $\times$ & $\times$ & LS$_\text{vid}$ & 17.4 & 13.9 & 13.0 & 17.1 & 21.4 & 11.9 & 7.0 & 31.6 & 15.5 & 14.4 & 14.9 & 17.8 & 53.0 \\
ChronoDepth~\cite{shao24learning} & $\times$ & $\times$ & LS$_\text{vid}$ & 15.3 & 14.8 & 14.7 & 14.5 & 21.1 & 11.3 & 7.0 & 31.0 & 14.3 & 16.8 & 14.5 & 17.7 & 7.0 \\
DepthCrafter~\cite{hu2024depthcrafter} & $\times$ & $\times$ & LS$_\text{vid}$ & 10.2 & 22.2 & 6.6 & 34.6 & 15.4 & 14.8 & 4.5 & 46.9 & 12.9 & 18.4 & 9.9 & 27.4 & 38.1 \\
DepthAny.~\cite{yang24depth}~($\ast$) & $\times$ & $\times$ & LS$_\text{vid}$  & 8.4 & 28.1 & 5.2 & 44.3 & 13.9 & 18.6 & 3.7 & 55.1 & 10.4 & 25.3 & 8.3 & 34.3 & \textbf{0.1} \\
DepthAny.V2~\cite{yang24depthv2} & $\times$ & $\times$ & LS$_\text{vid}$  & 9.1 & 26.3 & 4.6 & 50.0 & 16.1 & 16.0 & \textbf{3.2} & 61.6 & 10.1 & 26.5 & 8.6 & 36.1 & \textbf{0.1} \\
DUSt3R~\cite{wang24dust3r:} & $\times$ & $\times$ & LS$_\text{vid}$ & {7.2} & {31.7} & 4.6 & 50.0 & 13.4 & 19.0 & 3.4 & 59.2 & 9.3 & 28.4 & 7.6 & 37.7 & 4.3 \\
UniDepth~\cite{piccinelli24unidepth:} & $\times$ & $\times$ & LS$_\text{vid}$ & (3.9) & (58.2) & (1.5) & (90.1) & 12.7 & 18.9 & 7.0 & 30.7 & 10.6 & 23.2 & 7.1 & 44.2 & 0.5 \\
Ours & $\times$ & $\times$ & LS$_\text{vid}$ & \textbf{6.9} & \textbf{32.4} & \textbf{4.5} & \textbf{50.8} & \textbf{9.2} & \textbf{27.1} & \textbf{3.2} & \textbf{62.4} & \textbf{7.3} & \textbf{32.9} & \textbf{6.2} & \textbf{41.1} & 0.9$\ddag$ \\
\bottomrule
\end{tabular}
\vspace{-1em}
\caption{
\textbf{RobustMVD~\cite{schroppel2022benchmark} multi-view depth benchmark}, analogous to \cref{tab:depth_depthcrafter}.
Parentheses indicate that a model was trained on the same benchmark used for evaluation, which provides an advantage compared to training on different data.
$\ddag$Includes approximately 0.4 seconds for tracking.
$^\sharp$Metrics computed only for pixels with a valid prediction, which is an easier task.}%
\label{tab:depth_robustmvd}
\vspace{-1.5em}
\end{table*}

\begin{figure*}[ht]
\centering
\includegraphics[width=0.97\textwidth]{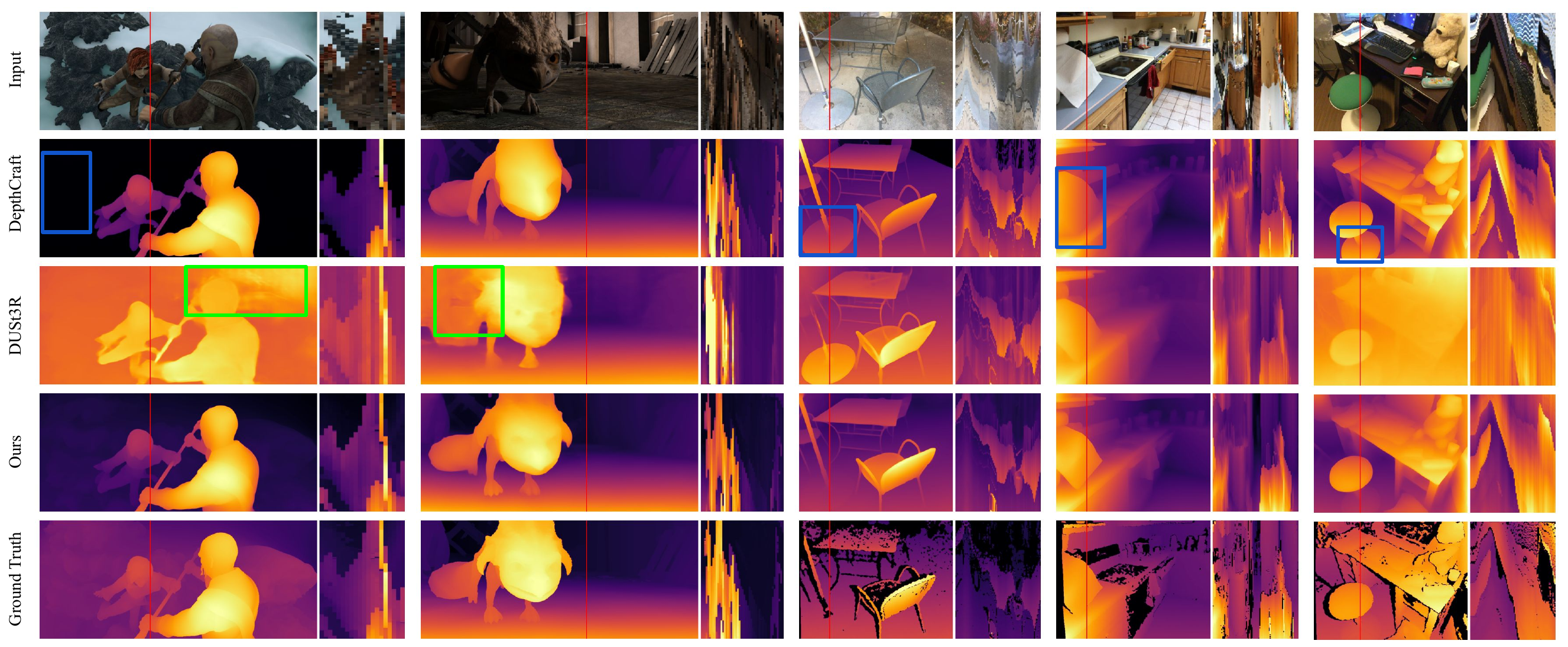}
\vspace{-1em}
\caption{\textbf{Video depth prediction}, comparing \mname (+DepthAnything), DepthCrafter~\cite{hu2024depthcrafter}, and DUSt3R~\cite{wang24dust3r:}.
We visualize a column of pixels (highlighted in red) over time to illustrate temporal variation.
Our model shows stable, coherent depth estimation over time, while DepthCrafter exhibits significant errors in certain regions (blue box).
DUSt3R struggles with dynamic content (green box).
}%
\label{fig:fig_depth}
\end{figure*}

\begin{figure*}[ht]
\centering
\vspace{-0.5em}
\includegraphics[width=0.97\textwidth,clip,trim={0 0 0 1em}]{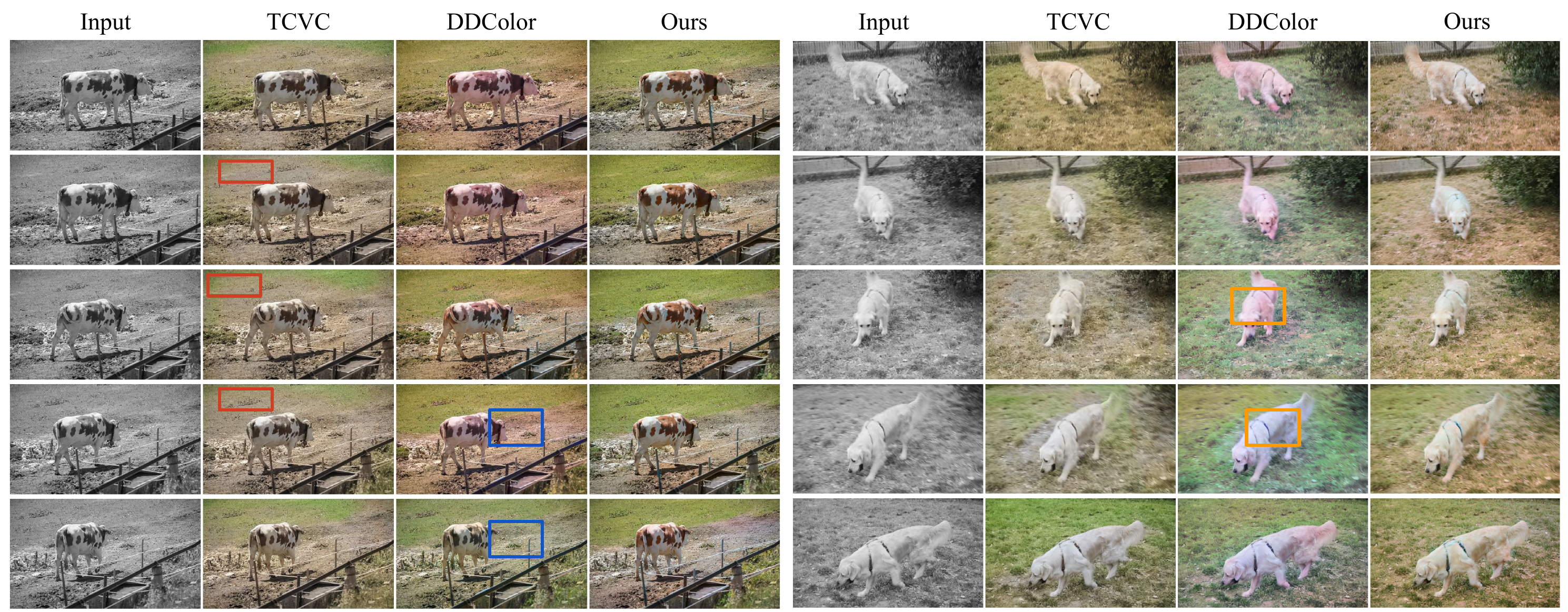}
\vspace{-1em}
\caption{\textbf{Qualitative comparison of video colorization methods.}
\mname (+DDColor) yields vibrant, realistic, and consistent colors. In contrast, TCVC appears less vibrant (21.7 vs. 29.5 Colorfulness), while DDColor lacks temporal consistency.}%
\label{fig:fig_colorization}
\vspace{-1em}
\end{figure*}

\begin{figure*}[ht]
    \centering
    \begin{minipage}[t]{0.65\textwidth}
    \vspace{0pt}
        \centering
        \small
        \tablestyle{3pt}{1.0}
        \begin{tabular}{l cccc cccc}
            \toprule
            \textbf{Method} & \multicolumn{4}{c}{\textbf{DAVIS (medium frame length)}} & \multicolumn{4}{c}{\textbf{Videvo (long frame length)}} \\
            \cmidrule(lr){2-5} \cmidrule(lr){6-9}
            & FID$\downarrow$ & CF$\uparrow$ & CDC$\downarrow$ & PSNR$\uparrow$ & FID$\downarrow$ & CF$\uparrow$ & CDC$\downarrow$ & PSNR$\uparrow$ \\
            \midrule
            \multicolumn{9}{l}{\emph{Specialized video colorization models:}} \\
            FAVC~\cite{lei2019fully} & --- & 18.55 & 4.22 & 24.38 & --- & 16.28 & 1.88 & 24.81 \\
            TCVC~\cite{liu2021temporally} & 46.51 & 21.70 & {3.73} & {25.50} & 39.58 & 19.07 & 1.64 & 25.43 \\
            \midrule 
            CIC~\cite{zhang2016colorful}   & 44.23 & 30.34 & 6.18 & \textbf{23.19} & 39.78 & 29.19 & 3.59 & 22.51 \\
            {}\cite{zhang2016colorful}+Ours & \textbf{43.13} & \textbf{34.22} & \textbf{4.77} \textcolor{blue}{(-22.8\%)} & {21.58} & \textbf{37.00} & \textbf{32.73} & \textbf{2.57} \textcolor{blue}{(-28.3\%)} & \textbf{22.64} \\
            \midrule 
            IDC~\cite{zhang2017real}   & 42.99  & 21.70 & 5.01 & \textbf{25.42} & 37.25 & 19.07 & 2.57 & \textbf{25.35} \\
            {}\cite{zhang2017real}+Ours & \textbf{33.47}  & \textbf{28.10}  & \textbf{4.45} \textcolor{blue}{(-11.1\%)} & 23.51 & \textbf{33.13} & \textbf{25.22} & \textbf{2.42} \textcolor{blue}{(-6.0\%)} & 24.04 \\
            \midrule 
            ColorFormer~\cite{ji2022colorformer}   & 40.71 & 29.61 & 8.29 & 23.03 & 40.12 & \textbf{29.76} & 4.96 & 23.08 \\
            {}\cite{ji2022colorformer}+Ours &  \textbf{29.73} & \textbf{32.24}  & \textbf{4.86} \textcolor{blue}{(-41.3\%)} & \textbf{23.18} & \textbf{27.25} & 29.65  & \textbf{3.53} \textcolor{blue}{(-28.8\%)} & \textbf{23.31}   \\
            \midrule 
            DDColor~\cite{kang2023ddcolor} & {26.81} & \textbf{30.61} & 8.64 & 23.81 & \textbf{20.27} & \textbf{30.38} & 5.60 & 24.34 \\
            {}\cite{kang2023ddcolor}+Ours & \textbf{24.61}  & 29.53 &  \textbf{4.62} \textcolor{blue}{(-46.5\%)} & \textbf{23.85} & 22.78 & 26.06 & \textbf{3.04} \textcolor{blue}{(-45.7\%)} & \textbf{24.39}  \\
            \bottomrule
        \end{tabular}\vspace{-1em}
        \captionof{table}{\textbf{Quantitative comparison of video colorization methods on the DAVIS and Videvo datasets.} Our method, when augmented onto four different baseline models, consistently improves the Color Distribution Consistency (CDC) metric across both datasets.}%
        \label{tab:colorization_main}
    \end{minipage}\hfill
    \begin{minipage}[t]{0.33\textwidth}
    \vspace{0pt}
        \centering
        \includegraphics[width=0.99\linewidth]{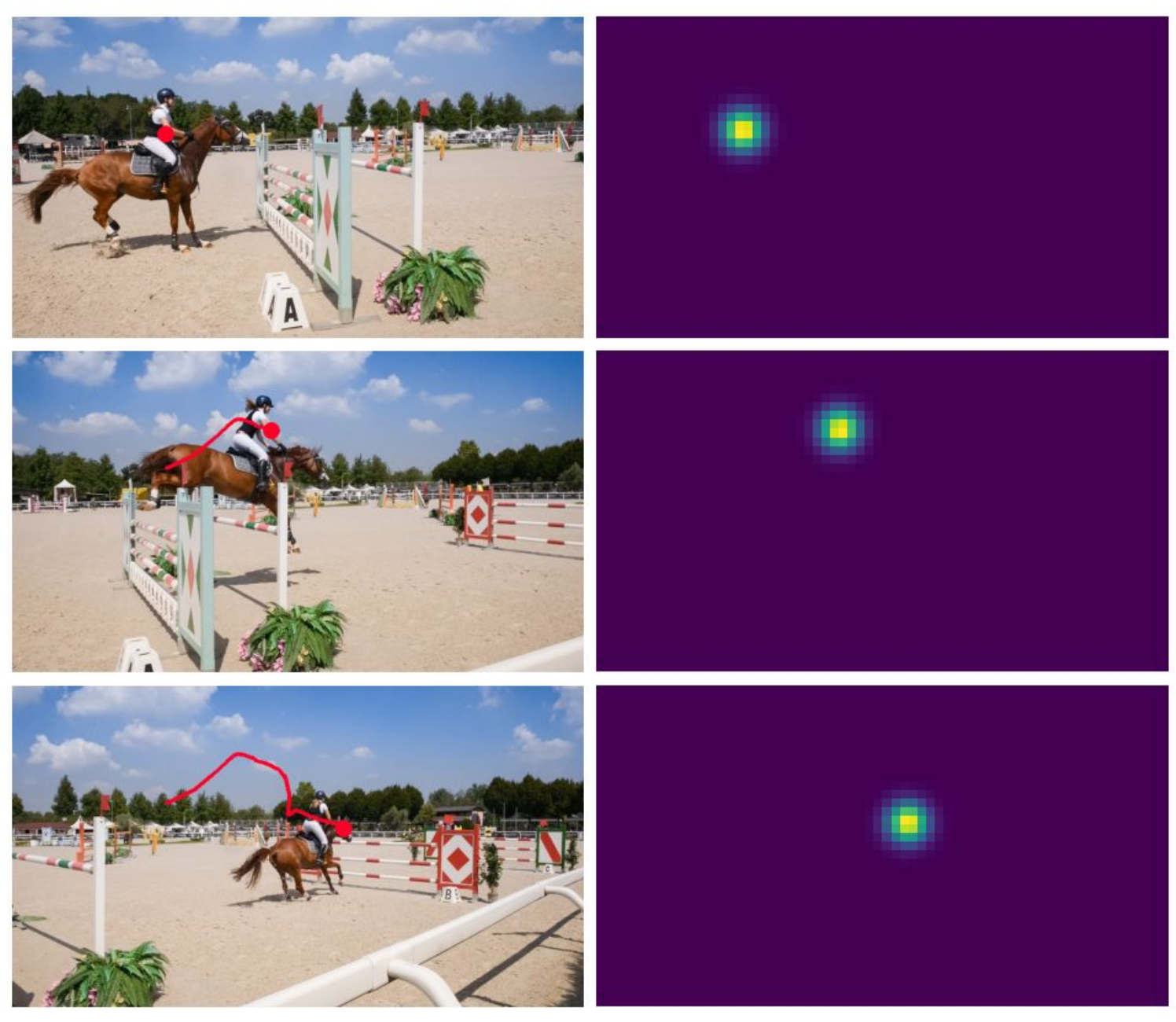}
        \caption{\textbf{Attentional Sampling module.} The module attends to image features in correspondence with each track, as shown by the attention maps next to each frame.}%
        \label{fig:fig_attn}
    \end{minipage}
\end{figure*}

\begin{table*}[t]
\centering
\vspace{-2mm}

\subfloat[\textbf{Number of training tracks}: Increasing training tracks consistently improved depth estimation accuracy, with optimal results at $24^2$ tracks.\label{tab:ablation:num_train_tracks}]{
\tablestyle{5pt}{1}
\begin{tabular}{x{38}x{25}x{25}x{25}}\toprule
\multicolumn{1}{c}{\#tracks}  & Sintel$\downarrow$ & ScanNet$\downarrow$ & T\&T$\downarrow$ \\
\midrule
$0$ (base) & 0.325 & 0.130 & 0.104  \\
$1$ & 0.347 & 0.094 & 0.087 \\
$12^2$ & 0.319 & \textbf{0.086} & 0.080 \\
\rowcolor[gray]{0.9}
$24^2$ & \textbf{0.295}  & 0.087  & \textbf{0.073}  \\
\bottomrule
\end{tabular}}%
\hspace{3mm}
\subfloat[\textbf{Number of testing tracks}: Increasing testing tracks significantly boosted performance, indicating better temporal capture with more tracks.\label{tab:ablation:num_test_tracks}]{
\tablestyle{5pt}{1}
\begin{tabular}{x{38}x{25}x{25}x{25}}\toprule
\multicolumn{1}{c}{\#tracks}  & Sintel$\downarrow$ & ScanNet$\downarrow$ & T\&T$\downarrow$ \\
\midrule
$0$ (base) & 0.325 & 0.130 & 0.104  \\
$1$ &  0.318 & 0.118 & 0.108 \\
$12^2$ & 0.321 &  \textbf{0.087} & {0.075} \\
\rowcolor[gray]{0.9}
$24^2$ & 0.295  & \textbf{0.087}  & \textbf{0.073}  \\
\bottomrule
\end{tabular}}%
\hspace{3mm}
\vspace{1mm}
\subfloat[\textbf{Type of tracks}: Using a point tracker performs better than using static pseudo-tracks.
Furthermore, querying tracks uniformly is better.\label{tab:ablation:track_type}]{
\tablestyle{5pt}{1}
\begin{tabular}{x{38}x{25}x{25}x{25}}\toprule
\multicolumn{1}{c}{Type}  & Sintel$\downarrow$ & ScanNet$\downarrow$ & T\&T$\downarrow$ \\
\midrule
Constant & 0.352  & 0.099 & 0.085  \\
Grid, $T_0$ & 0.325 & 0.096 & \textbf{0.073}  \\
Rand., $T_0$ & 0.320 & 0.095 & \textbf{0.073} \\
\rowcolor[gray]{0.9}
Rand., all & \textbf{0.295} & \textbf{0.087}  & \textbf{0.073}  \\
\bottomrule
\end{tabular}}%
\hspace{3mm}
\subfloat[\textbf{Temporal modeling}: \mname{} outperformed 3D conv.~and time attention by leveraging tracking data for better temporal alignment.\label{tab:ablation:temporal_modeling}]{
\tablestyle{5pt}{1}
\begin{tabular}{x{38}x{25}x{25}x{25}}\toprule
\multicolumn{1}{c}{Method}  & Sintel$\downarrow$ & ScanNet$\downarrow$ & T\&T$\downarrow$ \\
\midrule
    3D Conv. & 0.364 & 0.112 & 0.096 \\
    Time Attn. & 0.363 & 0.097 & 0.081 \\
    \rowcolor[gray]{0.9} %
     \mname{} & \textbf{0.295} & \textbf{0.087}  & \textbf{0.073}  \\
     \bottomrule
\end{tabular}}%
\hspace{3mm}
\subfloat[\textbf{Number of Time Attention Layers}: Two layers balanced performance and complexity, while additional layers offered no consistent benefits.\label{tab:ablation:attention_layers}]{
\tablestyle{5pt}{1}
\begin{tabular}{x{38}x{25}x{25}x{25}}\toprule
\multicolumn{1}{c}{\#Layers}  & Sintel$\downarrow$ & ScanNet$\downarrow$ & T\&T$\downarrow$ \\
\midrule
1 & 0.316 & 0.089 & 0.074 \\
\rowcolor[gray]{0.9}
2 & \textbf{0.295} & {0.087}  & \textbf{0.073}  \\
3 & 0.317 & \textbf{0.083} & \textbf{0.073}  \\
\bottomrule
\end{tabular}}%
\hspace{3mm}
\subfloat[\textbf{Number of \mname{} layers}: 6 layers improved temporal representation, but further layers led to diminishing returns due to overfitting risks.\label{tab:ablation:layer_pos}]{
\tablestyle{5pt}{1}
\begin{tabular}{x{38}x{25}x{25}x{25}}\toprule
\multicolumn{1}{c}{\#Layers}  & Sintel$\downarrow$ & ScanNet$\downarrow$ & T\&T$\downarrow$ \\
\midrule
1 & 0.344 & 0.090  & 0.087 \\
\rowcolor[gray]{0.9}
6 & \textbf{0.295} & \textbf{0.087}  & \textbf{0.073}  \\
12 & 0.333 & 0.105 & 0.075 \\
\bottomrule
\end{tabular}}%
\hspace{3mm}
\vspace{-0.1in}
\caption{\textbf{Ablations on video depth estimation results.} Each table examines the impact of different factors, such as the number and type of tracks used during training and testing. Results are reported using the AbsRel error metric, where $\downarrow$ indicates that lower values are better.}%
\label{tab:ablations}
\vspace{-0.2in}
\end{table*}

We show that \mname can turn image-based models into state-of-the-art ones for video depth estimation and colorization (\cref{sec:exp-depth,sec:exp-color}) and ablate its design (\cref{sec:ablation}).

\subsection{Video Depth Estimation}%
\label{sec:exp-depth}

To apply \mname to video depth estimation, we start from an image-based model, Depth Anything~\cite{yang2024depth}, and upgrade it to a video-based model using the \mname module.
The Depth Anything model has a DINO ViT backbone consisting of 12 transformer blocks.
We insert our \mname module after each of the last 6 blocks.

We fine-tune the model on a large dataset of videos, using a combination of synthetic and real data.
This data comes with ground-truth depth, which we fit using a scale- and shift-invariant loss, as Depth Anything predicts depth up to an affine transformation.
However, we share the \emph{same} calibration parameters for \emph{all} frames in a video, as calibrating frames independently would mask temporal inconsistencies, which we want the model to learn to correct.

To reduce the number of new parameters added to the model, we share them between all \mname modules.
During training, the original model is kept frozen, and only the \mname modules are updated.
We use the AdamW optimizer with an initial learning rate of $1.6 \times 10^{-5}$ and cosine learning rate decay.
Training is conducted for 4 epochs with a batch size of 4 videos, each containing a randomly sampled number of frames between 8 and 16.

\paragraph{Data and evaluation metrics.}

For training, we use a combination of datasets containing both synthetic and real videos. We use standard depth estimation metrics~\cite{yang24depth}. Please refer to Appendix~\ref{app:depth_imp_details} for details.

\subsubsection{Quantitative Results}

In \cref{tab:depth_depthcrafter}, we evaluate our model on four video collections in the DepthCrafter benchmark~\cite{hu2024depthcrafter} against several alternatives.
\mname improves the image-based model it augments, \ie, Depth Anything~\cite{yang24depth}, substantially:
Sintel AbsRel improves by 9.2\% and $\delta_1$ by 13.5\%,
KITTI by 26.8\% and 12.5\%,
Bonn by 15.4\% and 3.4\%,
and on average by 18.3\% and 9.7\%.
Second, \mname performs better than all other baselines, including DepthCrafter~\cite{hu2024depthcrafter}, which is a video depth model, by 15.3\% and 2.3\% on AbsRel and $\delta_{1.25}$, respectively.
This demonstrates that \mname can turn an image-based model into a video-based one, outperforming models designed for video data from the start.

\Cref{tab:depth_robustmvd} reports results on the RobustMVD benchmark~\cite{schroppel2022benchmark}, which features shorter video clips with potentially larger camera motions.
Our model achieved the lowest average relative error of 6.2 and the highest average inlier ratio of 41.1\%, surpassing all baselines.
In particular, it improves the base model Depth Anything by 33.8\% on ETH3D and by 29.8\% on Tanks and Temples.

Our model is also efficient, utilizing only 140 million parameters --- only 17.1 million for the \mname{} module.
It operates with a runtime of only 0.9 seconds per sequence, of which 0.4 seconds are attributed to the tracker.
Despite its smaller size, our model outperforms DepthCrafter~\cite{hu2024depthcrafter}, which relies on video diffusion models with over 1.5 billion parameters.
\mname will further benefit from future accuracy and speed improvements in point trackers.

\subsubsection{Qualitative Results}%
\label{sec:exp-depth-qualitative}

\Cref{fig:fig_depth} shows a qualitative comparison of depth estimation by our model, DepthCrafter, and DUSt3R in dynamic scenes, complex textures, and cluttered environments.
Our model produces stable, coherent depth maps across all frames, demonstrating strong temporal consistency.
In contrast, DepthCrafter shows significant depth estimation errors in certain regions (blue box), especially in scenes with complex structures.
DUSt3R, relying on implicit triangulation, struggles with dynamic content, resulting in inconsistent depth maps across time.

\Cref{fig:fig_attn} illustrates the Attentional Sampling block, which pools information along the track by attending to image features in correspondence with the track locations.

\subsection{Automatic Video Colorization}%
\label{sec:exp-color}

The goal of video colorization is to add colors to an input grayscale video.
Colors should be realistic, vibrant, and consistent across frames to avoid flickering.

We enhance four models --- CIC~\cite{zhang2016colorful}, IDC~\cite{zhang2017real}, Colorformer~\cite{ji2022colorformer}, and DDColor~\cite{kang2023ddcolor} --- by incorporating our \mname module into layers of each model’s architecture.
For instance, for DDColor, which has a ConvNet-based architecture with a ConvNeXt backbone~\cite{liu2022convnet} and a U-Net decoder, we insert our \mname module before each decoder block.
We fine-tune the model using AdamW on the YouTube-VIS training set~\cite{Yang2019vis}.
For testing, we use the validation sets of both DAVIS~\cite{perazzi2016benchmark} and Videvo~\cite{lai2018learning}.

\paragraph{Evaluation metrics.} We evaluate colorization results using standard metrics. Please refer to Appendix~\ref{app:color_imp_details} for details.

\paragraph{Quantitative Results.}

\Cref{fig:fig_colorization} reports video colorization results on DAVIS and Videvo.
We apply \mname to several image colorization models, consistently improving their performance when applied to videos, particularly in color consistency.
Specifically, CDC improves by
22.8\% on DAVIS and 28.3\% on Videvo for CIC~\cite{zhang2016colorful},
41.3\% and 28.8\% for Colorformer~\cite{ji2022colorformer}, and
46.5\% and 45.7\% for DDColor~\cite{kang2023ddcolor}.
\mname also improves FID and Colorfulness, indicating overall improvements in visual quality.
We also compare to native video colorization networks (top rows), often outperforming these as well.

\paragraph{Qualitative Results.}

In \cref{fig:fig_colorization}, our method delivers more vibrant, realistic, and consistent colors compared to TCVC and DDColor.
TCVC's colors are less vibrant, with a colorfulness score of 21.7 compared to our 29.5, and DDColor exhibits inconsistencies across frames, with colors sometimes changing dramatically over time.

\subsection{Ablation Studies}%
\label{sec:ablation}

Here, we measure the impact of various design decisions in \mname using the AbsRel metric on depth prediction in Sintel, ScanNet, and Tanks and Temples.

\Cref{tab:ablation:num_train_tracks} assesses the impact of the \emph{number of tracks} used in \emph{training}:
even a single track per video can lead to improvements, and the optimal number is around $24^2$ tracks.
\Cref{tab:ablation:num_test_tracks} assesses the \emph{number of tracks} used at \emph{testing} time, with similar results.
\Cref{tab:ablation:track_type} compares different tracking strategies:
eschewing the tracker and simply setting tracks to be constant across all frames,
and initializing the tracks on a grid in the first frame, at random in the first frame, or at random throughout the video.
The latter results in much better performance, showing the importance of tracking and ensuring that tracks cover the video well.
\Cref{tab:ablation:temporal_modeling} tests replacing \mname Layers with 3D convolutions and time attention, which, by comparison, barely improve results.
\Cref{tab:ablation:attention_layers} shows the impact of increasing the \emph{number of transformer layers} in the Track Transformer, where two layers are optimal in most cases.
Finally, \cref{tab:ablation:layer_pos} shows the impact of adding a different \emph{number of \mname layers}, concluding that six is about optimal.

\subsection{Limitations}

\mname's success depends on the quality of the extracted tracks.
While modern point trackers are robust and efficient, they may still fail, particularly when tracks are off-camera for extended periods.
Furthermore, running a point tracker in addition to the primary neural network adds complexity and computational cost (although this is often offset by avoiding spatio-temporal attention).

\section{Conclusion}%
\label{sec:conclusion}

We have introduced \mname, a plug-and-play layer that upgrades image predictor networks into video predictor ones.
It leverages the power of modern point trackers, transferring their understanding of motion to new applications such as depth prediction and colorization.
The module is lightweight, simple to integrate, and compatible with a variety of architectures, including Vision Transformers and ConvNets.
Thanks to the efficiency of recent point trackers like CoTracker3, it incurs only a small computational overhead.
Remarkably, the resulting upgraded image models are \emph{overall} faster and more accurate than some state-of-the-art methods specifically designed for video prediction.

\paragraph*{Acknowledgments.}

This research was supported by ERC 101001212-UNION\@.

{
\small
\bibliographystyle{ieeenat_fullname}
\bibliography{main,vedaldi_general,vedaldi_specific}

\begin{thebibliography}{69}
\providecommand{\natexlab}[1]{#1}
\providecommand{\url}[1]{\texttt{#1}}
\expandafter\ifx\csname urlstyle\endcsname\relax
  \providecommand{\doi}[1]{doi: #1}\else
  \providecommand{\doi}{doi: \begingroup \urlstyle{rm}\Url}\fi

\bibitem[Arnab et~al.(2021)Arnab, Dehghani, Heigold, Sun, Lu{\v{c}}i{\'c}, and Schmid]{arnab2021vivit}
Anurag Arnab, Mostafa Dehghani, Georg Heigold, Chen Sun, Mario Lu{\v{c}}i{\'c}, and Cordelia Schmid.
\newblock Vivit: A video vision transformer.
\newblock In \emph{Proceedings of the IEEE/CVF international conference on computer vision}, pages 6836--6846, 2021.

\bibitem[Baruch et~al.(2021)Baruch, Chen, Dehghan, Dimry, Feigin, Fu, Gebauer, Joffe, Kurz, Schwartz, et~al.]{baruch2021arkitscenes}
Gilad Baruch, Zhuoyuan Chen, Afshin Dehghan, Tal Dimry, Yuri Feigin, Peter Fu, Thomas Gebauer, Brandon Joffe, Daniel Kurz, Arik Schwartz, et~al.
\newblock Arkitscenes: A diverse real-world dataset for 3d indoor scene understanding using mobile rgb-d data.
\newblock \emph{arXiv preprint arXiv:2111.08897}, 2021.

\bibitem[Bertasius et~al.(2021)Bertasius, Wang, and Torresani]{bertasius2021space}
Gedas Bertasius, Heng Wang, and Lorenzo Torresani.
\newblock Is space-time attention all you need for video understanding?
\newblock In \emph{ICML}, page~4, 2021.

\bibitem[Blattmann et~al.(2023{\natexlab{a}})Blattmann, Dockhorn, Kulal, Mendelevitch, Kilian, Lorenz, Levi, English, Voleti, Letts, et~al.]{blattmann2023stable}
Andreas Blattmann, Tim Dockhorn, Sumith Kulal, Daniel Mendelevitch, Maciej Kilian, Dominik Lorenz, Yam Levi, Zion English, Vikram Voleti, Adam Letts, et~al.
\newblock Stable video diffusion: Scaling latent video diffusion models to large datasets.
\newblock \emph{arXiv preprint arXiv:2311.15127}, 2023{\natexlab{a}}.

\bibitem[Blattmann et~al.(2023{\natexlab{b}})Blattmann, Rombach, Ling, Dockhorn, Kim, Fidler, and Kreis]{blattmann2023align}
Andreas Blattmann, Robin Rombach, Huan Ling, Tim Dockhorn, Seung~Wook Kim, Sanja Fidler, and Karsten Kreis.
\newblock Align your latents: High-resolution video synthesis with latent diffusion models.
\newblock In \emph{Proceedings of the IEEE/CVF Conference on Computer Vision and Pattern Recognition}, pages 22563--22575, 2023{\natexlab{b}}.

\bibitem[Carreira and Zisserman(2017)]{carreira2017quo}
Joao Carreira and Andrew Zisserman.
\newblock Quo vadis, action recognition? a new model and the kinetics dataset.
\newblock In \emph{proceedings of the IEEE Conference on Computer Vision and Pattern Recognition}, pages 6299--6308, 2017.

\bibitem[Cho et~al.(2024)Cho, Huang, Nam, An, Kim, and Lee]{cho2024local}
Seokju Cho, Jiahui Huang, Jisu Nam, Honggyu An, Seungryong Kim, and Joon-Young Lee.
\newblock Local all-pair correspondence for point tracking.
\newblock \emph{arXiv preprint arXiv:2407.15420}, 2024.

\bibitem[Dai et~al.(2017)Dai, Chang, Savva, Halber, Funkhouser, and Nie{\ss}ner]{dai2017scannet}
Angela Dai, Angel~X Chang, Manolis Savva, Maciej Halber, Thomas Funkhouser, and Matthias Nie{\ss}ner.
\newblock Scannet: Richly-annotated 3d reconstructions of indoor scenes.
\newblock In \emph{Proceedings of the IEEE conference on computer vision and pattern recognition}, pages 5828--5839, 2017.

\bibitem[Dehghani et~al.(2023)Dehghani, Djolonga, Mustafa, Padlewski, Heek, Gilmer, Steiner, Caron, Geirhos, Alabdulmohsin, et~al.]{dehghani2023scaling}
Mostafa Dehghani, Josip Djolonga, Basil Mustafa, Piotr Padlewski, Jonathan Heek, Justin Gilmer, Andreas~Peter Steiner, Mathilde Caron, Robert Geirhos, Ibrahim Alabdulmohsin, et~al.
\newblock Scaling vision transformers to 22 billion parameters.
\newblock In \emph{International Conference on Machine Learning}, pages 7480--7512. PMLR, 2023.

\bibitem[Doersch et~al.(2022)Doersch, Gupta, Markeeva, Recasens, Smaira, Aytar, Carreira, Zisserman, and Yang]{doersch2022tap}
Carl Doersch, Ankush Gupta, Larisa Markeeva, Adria Recasens, Lucas Smaira, Yusuf Aytar, Joao Carreira, Andrew Zisserman, and Yi Yang.
\newblock Tap-vid: A benchmark for tracking any point in a video.
\newblock \emph{Advances in Neural Information Processing Systems}, 35:\penalty0 13610--13626, 2022.

\bibitem[Doersch et~al.(2023{\natexlab{a}})Doersch, Yang, Vecerik, Gokay, Gupta, Aytar, Carreira, and Zisserman]{doersch2023tapir}
Carl Doersch, Yi Yang, Mel Vecerik, Dilara Gokay, Ankush Gupta, Yusuf Aytar, Joao Carreira, and Andrew Zisserman.
\newblock Tapir: Tracking any point with per-frame initialization and temporal refinement.
\newblock In \emph{Proceedings of the IEEE/CVF International Conference on Computer Vision}, pages 10061--10072, 2023{\natexlab{a}}.

\bibitem[Doersch et~al.(2023{\natexlab{b}})Doersch, Yang, Vecerik, Gokay, Gupta, Aytar, Carreira, and Zisserman]{doersch23tapir:}
Carl Doersch, Yi Yang, Mel Vecerik, Dilara Gokay, Ankush Gupta, Yusuf Aytar, Joao Carreira, and Andrew Zisserman.
\newblock {TAPIR:} tracking any point with per-frame initialization and temporal refinement.
\newblock In \emph{Proc. {CVPR}}, 2023{\natexlab{b}}.

\bibitem[Doersch et~al.(2024)Doersch, Yang, Gokay, Luc, Koppula, Gupta, Heyward, Goroshin, Carreira, and Zisserman]{doersch24bootstap:}
Carl Doersch, Yi Yang, Dilara Gokay, Pauline Luc, Skanda Koppula, Ankush Gupta, Joseph Heyward, Ross Goroshin, Jo{\~a}o Carreira, and Andrew Zisserman.
\newblock {BootsTAP}: Bootstrapped training for tracking-any-point.
\newblock \emph{arXiv}, 2402.00847, 2024.

\bibitem[Fan et~al.(2021{\natexlab{a}})Fan, Xiong, Mangalam, Li, Yan, and Malik{\ldots}]{fan21multiscale}
H Fan, B Xiong, K Mangalam, Y Li, Z Yan, and J Malik{\ldots}.
\newblock Multiscale vision transformers.
\newblock In \emph{Proc. {CVPR}}, 2021{\natexlab{a}}.

\bibitem[Fan et~al.(2021{\natexlab{b}})Fan, Xiong, Mangalam, Li, Yan, Malik, and Feichtenhofer]{fan2021multiscale}
Haoqi Fan, Bo Xiong, Karttikeya Mangalam, Yanghao Li, Zhicheng Yan, Jitendra Malik, and Christoph Feichtenhofer.
\newblock Multiscale vision transformers.
\newblock In \emph{Proceedings of the IEEE/CVF international conference on computer vision}, pages 6824--6835, 2021{\natexlab{b}}.

\bibitem[Feichtenhofer et~al.(2019)Feichtenhofer, Fan, Malik, and He]{feichtenhofer2019slowfast}
Christoph Feichtenhofer, Haoqi Fan, Jitendra Malik, and Kaiming He.
\newblock Slowfast networks for video recognition.
\newblock In \emph{Proceedings of the IEEE/CVF international conference on computer vision}, pages 6202--6211, 2019.

\bibitem[Geiger et~al.(2013)Geiger, Lenz, Stiller, and Urtasun]{geiger2013vision}
Andreas Geiger, Philip Lenz, Christoph Stiller, and Raquel Urtasun.
\newblock Vision meets robotics: The kitti dataset.
\newblock \emph{The International Journal of Robotics Research}, 32\penalty0 (11):\penalty0 1231--1237, 2013.

\bibitem[Harley et~al.(2022{\natexlab{a}})Harley, Fang, and Fragkiadaki]{harley2022particle}
Adam~W Harley, Zhaoyuan Fang, and Katerina Fragkiadaki.
\newblock Particle video revisited: Tracking through occlusions using point trajectories.
\newblock In \emph{European Conference on Computer Vision}, pages 59--75. Springer, 2022{\natexlab{a}}.

\bibitem[Harley et~al.(2022{\natexlab{b}})Harley, Fang, and Fragkiadaki]{harley22particle}
Adam~W. Harley, Zhaoyuan Fang, and Katerina Fragkiadaki.
\newblock Particle videos revisited: Tracking through occlusions using point trajectories.
\newblock In \emph{Proc. {ECCV}}, 2022{\natexlab{b}}.

\bibitem[Hu et~al.(2024)Hu, Gao, Li, Zhao, Cun, Zhang, Quan, and Shan]{hu2024depthcrafter}
Wenbo Hu, Xiangjun Gao, Xiaoyu Li, Sijie Zhao, Xiaodong Cun, Yong Zhang, Long Quan, and Ying Shan.
\newblock Depthcrafter: Generating consistent long depth sequences for open-world videos.
\newblock \emph{arXiv preprint arXiv:2409.02095}, 2024.

\bibitem[Ji et~al.(2022)Ji, Jiang, Luo, Tao, Chu, Xie, Wang, and Tai]{ji2022colorformer}
Xiaozhong Ji, Boyuan Jiang, Donghao Luo, Guangpin Tao, Wenqing Chu, Zhifeng Xie, Chengjie Wang, and Ying Tai.
\newblock Colorformer: Image colorization via color memory assisted hybrid-attention transformer.
\newblock In \emph{European Conference on Computer Vision}, pages 20--36. Springer, 2022.

\bibitem[Kang et~al.(2023)Kang, Yang, Ouyang, Ren, Li, and Xie]{kang2023ddcolor}
Xiaoyang Kang, Tao Yang, Wenqi Ouyang, Peiran Ren, Lingzhi Li, and Xuansong Xie.
\newblock Ddcolor: Towards photo-realistic image colorization via dual decoders.
\newblock In \emph{Proceedings of the IEEE/CVF International Conference on Computer Vision}, pages 328--338, 2023.

\bibitem[Karaev et~al.(2023)Karaev, Rocco, Graham, Neverova, Vedaldi, and Rupprecht]{karaev2023dynamicstereo}
Nikita Karaev, Ignacio Rocco, Benjamin Graham, Natalia Neverova, Andrea Vedaldi, and Christian Rupprecht.
\newblock Dynamicstereo: Consistent dynamic depth from stereo videos.
\newblock In \emph{Proceedings of the IEEE/CVF Conference on Computer Vision and Pattern Recognition}, pages 13229--13239, 2023.

\bibitem[Karaev et~al.(2024{\natexlab{a}})Karaev, Makarov, Wang, Neverova, Vedaldi, and Rupprecht]{karaev2024cotracker3}
Nikita Karaev, Iurii Makarov, Jianyuan Wang, Natalia Neverova, Andrea Vedaldi, and Christian Rupprecht.
\newblock {CoTracker3}: Simpler and better point tracking by pseudo-labelling real videos.
\newblock \emph{arxiv}, 2024{\natexlab{a}}.

\bibitem[Karaev et~al.(2024{\natexlab{b}})Karaev, Rocco, Graham, Neverova, Vedaldi, and Rupprecht]{karaev23cotracker}
Nikita Karaev, Ignacio Rocco, Benjamin Graham, Natalia Neverova, Andrea Vedaldi, and Christian Rupprecht.
\newblock Cotracker: It is better to track together.
\newblock In \emph{Proc. {ECCV}}, 2024{\natexlab{b}}.

\bibitem[Karaev et~al.(2024{\natexlab{c}})Karaev, Rocco, Graham, Neverova, Vedaldi, and Rupprecht]{karaev24cotracker}
Nikita Karaev, Ignacio Rocco, Ben Graham, Natalia Neverova, Andrea Vedaldi, and Christian Rupprecht.
\newblock {CoTracker}: It is better to track together.
\newblock In \emph{Proceedings of the European Conference on Computer Vision ({ECCV})}, 2024{\natexlab{c}}.

\bibitem[Ke et~al.(2024)Ke, Obukhov, Huang, Metzger, Daudt, and Schindler]{ke2024repurposing}
Bingxin Ke, Anton Obukhov, Shengyu Huang, Nando Metzger, Rodrigo~Caye Daudt, and Konrad Schindler.
\newblock Repurposing diffusion-based image generators for monocular depth estimation.
\newblock In \emph{Proceedings of the IEEE/CVF Conference on Computer Vision and Pattern Recognition}, pages 9492--9502, 2024.

\bibitem[Knapitsch et~al.(2017)Knapitsch, Park, Zhou, and Koltun]{knapitsch2017tanks}
Arno Knapitsch, Jaesik Park, Qian-Yi Zhou, and Vladlen Koltun.
\newblock Tanks and temples: Benchmarking large-scale scene reconstruction.
\newblock \emph{ACM Transactions on Graphics (ToG)}, 36\penalty0 (4):\penalty0 1--13, 2017.

\bibitem[Kopf et~al.(2021)Kopf, Rong, and Huang]{kopf2021robust}
Johannes Kopf, Xuejian Rong, and Jia-Bin Huang.
\newblock Robust consistent video depth estimation.
\newblock In \emph{Proceedings of the IEEE/CVF Conference on Computer Vision and Pattern Recognition}, pages 1611--1621, 2021.

\bibitem[Lai et~al.(2018)Lai, Huang, Wang, Shechtman, Yumer, and Yang]{lai2018learning}
Wei-Sheng Lai, Jia-Bin Huang, Oliver Wang, Eli Shechtman, Ersin Yumer, and Ming-Hsuan Yang.
\newblock Learning blind video temporal consistency.
\newblock In \emph{Proceedings of the European conference on computer vision (ECCV)}, pages 170--185, 2018.

\bibitem[Lei and Chen(2019)]{lei2019fully}
Chenyang Lei and Qifeng Chen.
\newblock Fully automatic video colorization with self-regularization and diversity.
\newblock In \emph{Proceedings of the IEEE/CVF conference on computer vision and pattern recognition}, pages 3753--3761, 2019.

\bibitem[Li et~al.(2025)Li, Zhang, Liu, Zeng, Ren, Li, and Zhang]{li2025taptr}
Hongyang Li, Hao Zhang, Shilong Liu, Zhaoyang Zeng, Tianhe Ren, Feng Li, and Lei Zhang.
\newblock Taptr: Tracking any point with transformers as detection.
\newblock In \emph{European Conference on Computer Vision}, pages 57--75. Springer, 2025.

\bibitem[Ling et~al.(2024)Ling, Sheng, Tu, Zhao, Xin, Wan, Yu, Guo, Yu, Lu, et~al.]{ling2024dl3dv}
Lu Ling, Yichen Sheng, Zhi Tu, Wentian Zhao, Cheng Xin, Kun Wan, Lantao Yu, Qianyu Guo, Zixun Yu, Yawen Lu, et~al.
\newblock Dl3dv-10k: A large-scale scene dataset for deep learning-based 3d vision.
\newblock In \emph{Proceedings of the IEEE/CVF Conference on Computer Vision and Pattern Recognition}, pages 22160--22169, 2024.

\bibitem[Liu et~al.(2021)Liu, Zhao, Chan, Wang, Loy, Qiao, and Dong]{liu2021temporally}
Yihao Liu, Hengyuan Zhao, Kelvin~CK Chan, Xintao Wang, Chen~Change Loy, Yu Qiao, and Chao Dong.
\newblock Temporally consistent video colorization with deep feature propagation and self-regularization learning.
\newblock \emph{arXiv preprint arXiv:2110.04562}, 2021.

\bibitem[Liu et~al.(2017)Liu, Yeh, Tang, Liu, and Agarwala]{liu2017video}
Ziwei Liu, Raymond~A Yeh, Xiaoou Tang, Yiming Liu, and Aseem Agarwala.
\newblock Video frame synthesis using deep voxel flow.
\newblock In \emph{Proceedings of the IEEE international conference on computer vision}, pages 4463--4471, 2017.

\bibitem[Liu et~al.(2022{\natexlab{a}})Liu, Mao, Wu, Feichtenhofer, Darrell, and Xie]{liu2022convnet}
Zhuang Liu, Hanzi Mao, Chao-Yuan Wu, Christoph Feichtenhofer, Trevor Darrell, and Saining Xie.
\newblock A convnet for the 2020s.
\newblock In \emph{Proceedings of the IEEE/CVF conference on computer vision and pattern recognition}, pages 11976--11986, 2022{\natexlab{a}}.

\bibitem[Liu et~al.(2022{\natexlab{b}})Liu, Ning, Cao, Wei, Zhang, Lin, and Hu]{liu2022video}
Ze Liu, Jia Ning, Yue Cao, Yixuan Wei, Zheng Zhang, Stephen Lin, and Han Hu.
\newblock Video swin transformer.
\newblock In \emph{Proceedings of the IEEE/CVF conference on computer vision and pattern recognition}, pages 3202--3211, 2022{\natexlab{b}}.

\bibitem[Luo et~al.(2020)Luo, Huang, Szeliski, Matzen, and Kopf]{luo2020consistent}
Xuan Luo, Jia-Bin Huang, Richard Szeliski, Kevin Matzen, and Johannes Kopf.
\newblock Consistent video depth estimation.
\newblock \emph{ACM Transactions on Graphics (ToG)}, 39\penalty0 (4):\penalty0 71--1, 2020.

\bibitem[Mayer et~al.(2016)Mayer, Ilg, Hausser, Fischer, Cremers, Dosovitskiy, and Brox]{mayer2016large}
Nikolaus Mayer, Eddy Ilg, Philip Hausser, Philipp Fischer, Daniel Cremers, Alexey Dosovitskiy, and Thomas Brox.
\newblock A large dataset to train convolutional networks for disparity, optical flow, and scene flow estimation.
\newblock In \emph{Proceedings of the IEEE conference on computer vision and pattern recognition}, pages 4040--4048, 2016.

\bibitem[Neimark et~al.(2021)Neimark, Bar, Zohar, and Asselmann]{neimark21video}
Daniel Neimark, Omri Bar, Maya Zohar, and Dotan Asselmann.
\newblock Video transformer network.
\newblock In \emph{Proc. {ICCV} Workshops}, 2021.

\bibitem[Palazzolo et~al.(2019)Palazzolo, Behley, Lottes, Gigu\`ere, and Stachniss]{palazzolo2019iros}
E. Palazzolo, J. Behley, P. Lottes, P. Gigu\`ere, and C. Stachniss.
\newblock {ReFusion: 3D Reconstruction in Dynamic Environments for RGB-D Cameras Exploiting Residuals}.
\newblock In \emph{2019 IEEE/RSJ International Conference on Intelligent Robots and Systems (IROS)}, 2019.

\bibitem[Patrick et~al.(2021)Patrick, Campbell, Asano, Misra, Metze, Feichtenhofer, Vedaldi, and Henriques]{patrick21keeping}
Mandela Patrick, Dylan Campbell, Yuki~Markus Asano, Ishan Misra, Florian Metze, Christoph Feichtenhofer, Andrea Vedaldi, and Jo{\~{a}}o~F. Henriques.
\newblock Keeping your eye on the ball: Trajectory attention in video transformers.
\newblock In \emph{Proceedings of Advances in Neural Information Processing Systems (NeurIPS)}, 2021.

\bibitem[Perazzi et~al.(2016)Perazzi, Pont-Tuset, McWilliams, Van~Gool, Gross, and Sorkine-Hornung]{perazzi2016benchmark}
Federico Perazzi, Jordi Pont-Tuset, Brian McWilliams, Luc Van~Gool, Markus Gross, and Alexander Sorkine-Hornung.
\newblock A benchmark dataset and evaluation methodology for video object segmentation.
\newblock In \emph{Proceedings of the IEEE conference on computer vision and pattern recognition}, pages 724--732, 2016.

\bibitem[Piccinelli et~al.(2024)Piccinelli, Yang, Sakaridis, Segu, Li, Gool, and Yu]{piccinelli24unidepth:}
Luigi Piccinelli, Yung-Hsu Yang, Christos Sakaridis, Mattia Segu, Siyuan Li, Luc~Van Gool, and Fisher Yu.
\newblock {UniDepth:} universal monocular metric depth estimation.
\newblock In \emph{Proc. {CVPR}}, 2024.

\bibitem[Sch\"{o}nberger and Frahm(2016)]{schonberger16structure-from-motion}
Johannes~Lutz Sch\"{o}nberger and Jan-Michael Frahm.
\newblock Structure-from-motion revisited.
\newblock In \emph{Proc. {CVPR}}, 2016.

\bibitem[Sch\"{o}nberger et~al.(2016)Sch\"{o}nberger, Zheng, Pollefeys, and Frahm]{schonberger16pixelwise}
Johannes~Lutz Sch\"{o}nberger, Enliang Zheng, Marc Pollefeys, and Jan-Michael Frahm.
\newblock Pixelwise view selection for unstructured multi-view stereo.
\newblock In \emph{Proc. {ECCV}}, 2016.

\bibitem[Schops et~al.(2017)Schops, Schonberger, Galliani, Sattler, Schindler, Pollefeys, and Geiger]{Schops_2017_CVPR}
Thomas Schops, Johannes~L. Schonberger, Silvano Galliani, Torsten Sattler, Konrad Schindler, Marc Pollefeys, and Andreas Geiger.
\newblock A multi-view stereo benchmark with high-resolution images and multi-camera videos.
\newblock In \emph{Proceedings of the IEEE Conference on Computer Vision and Pattern Recognition (CVPR)}, 2017.

\bibitem[Schr{\"o}ppel et~al.(2022)Schr{\"o}ppel, Bechtold, Amiranashvili, and Brox]{schroppel2022benchmark}
Philipp Schr{\"o}ppel, Jan Bechtold, Artemij Amiranashvili, and Thomas Brox.
\newblock A benchmark and a baseline for robust multi-view depth estimation.
\newblock In \emph{2022 International Conference on 3D Vision (3DV)}, pages 637--645. IEEE, 2022.

\bibitem[Shao et~al.(2024)Shao, Yang, Zhou, Zhang, Shen, Poggi, and Liao]{shao24learning}
Jiahao Shao, Yuanbo Yang, Hongyu Zhou, Youmin Zhang, Yujun Shen, Matteo Poggi, and Yiyi Liao.
\newblock Learning temporally consistent video depth from video diffusion priors.
\newblock \emph{arXiv}, 2406.01493, 2024.

\bibitem[Singer et~al.(2022)Singer, Polyak, Hayes, Yin, An, Zhang, Hu, Yang, Ashual, Gafni, et~al.]{singer2022make}
Uriel Singer, Adam Polyak, Thomas Hayes, Xi Yin, Jie An, Songyang Zhang, Qiyuan Hu, Harry Yang, Oron Ashual, Oran Gafni, et~al.
\newblock Make-a-video: Text-to-video generation without text-video data.
\newblock \emph{arXiv preprint arXiv:2209.14792}, 2022.

\bibitem[Su et~al.(2021)Su, Lu, Pan, Wen, and Liu]{su2021roformer}
Jianlin Su, Yu Lu, Shengfeng Pan, Bo Wen, and Yunfeng Liu.
\newblock Roformer: Enhanced transformer with rotary position embedding, 2021.

\bibitem[Teed and Deng(2020)]{teed2020raft}
Zachary Teed and Jia Deng.
\newblock Raft: Recurrent all-pairs field transforms for optical flow.
\newblock In \emph{Computer Vision--ECCV 2020: 16th European Conference, Glasgow, UK, August 23--28, 2020, Proceedings, Part II 16}, pages 402--419. Springer, 2020.

\bibitem[Tran et~al.(2015)Tran, Bourdev, Fergus, Torresani, and Paluri]{tran2015learning}
Du Tran, Lubomir Bourdev, Rob Fergus, Lorenzo Torresani, and Manohar Paluri.
\newblock Learning spatiotemporal features with 3d convolutional networks.
\newblock In \emph{Proceedings of the IEEE international conference on computer vision}, pages 4489--4497, 2015.

\bibitem[Wang et~al.(2023{\natexlab{a}})Wang, Hu, He, Wang, Yu, Tuytelaars, Xu, and Wu]{wang23neural}
Liao Wang, Qiang Hu, Qihan He, Ziyu Wang, Jingyi Yu, Tinne Tuytelaars, Lan Xu, and Minye Wu.
\newblock Neural residual radiance fields for streamably free-viewpoint videos.
\newblock In \emph{Proc. {CVPR}}, 2023{\natexlab{a}}.

\bibitem[Wang et~al.(2023{\natexlab{b}})Wang, Chang, Cai, Li, Hariharan, Holynski, and Snavely]{wang2023tracking}
Qianqian Wang, Yen-Yu Chang, Ruojin Cai, Zhengqi Li, Bharath Hariharan, Aleksander Holynski, and Noah Snavely.
\newblock Tracking everything everywhere all at once.
\newblock In \emph{Proceedings of the IEEE/CVF International Conference on Computer Vision}, pages 19795--19806, 2023{\natexlab{b}}.

\bibitem[Wang et~al.(2024)Wang, Leroy, Cabon, Chidlovskii, and Revaud]{wang24dust3r:}
Shuzhe Wang, Vincent Leroy, Yohann Cabon, Boris Chidlovskii, and Jerome Revaud.
\newblock {DUSt3R}: Geometric {3D} vision made easy.
\newblock In \emph{Proc. {CVPR}}, 2024.

\bibitem[Wang et~al.(2020)Wang, Zhu, Wang, Hu, Qiu, Wang, Hu, Kapoor, and Scherer]{wang2020tartanair}
Wenshan Wang, Delong Zhu, Xiangwei Wang, Yaoyu Hu, Yuheng Qiu, Chen Wang, Yafei Hu, Ashish Kapoor, and Sebastian Scherer.
\newblock Tartanair: A dataset to push the limits of visual slam.
\newblock In \emph{2020 IEEE/RSJ International Conference on Intelligent Robots and Systems (IROS)}, pages 4909--4916. IEEE, 2020.

\bibitem[Wang et~al.(2018)Wang, Girshick, Gupta, and He]{wang18non-local}
Xiaolong Wang, Ross~B. Girshick, Abhinav Gupta, and Kaiming He.
\newblock Non-local neural networks.
\newblock In \emph{Proc. {CVPR}}, 2018.

\bibitem[Wang et~al.(2023{\natexlab{c}})Wang, Shi, Li, Huang, Cao, Zhang, Xian, and Lin]{wang2023neural}
Yiran Wang, Min Shi, Jiaqi Li, Zihao Huang, Zhiguo Cao, Jianming Zhang, Ke Xian, and Guosheng Lin.
\newblock Neural video depth stabilizer.
\newblock In \emph{Proceedings of the IEEE/CVF International Conference on Computer Vision}, pages 9466--9476, 2023{\natexlab{c}}.

\bibitem[Yang et~al.(2019)Yang, Fan, and Xu]{Yang2019vis}
Linjie Yang, Yuchen Fan, and Ning Xu.
\newblock Video instance segmentation.
\newblock In \emph{ICCV}, 2019.

\bibitem[Yang et~al.(2024{\natexlab{a}})Yang, Kang, Huang, Xu, Feng, and Zhao]{yang2024depth}
Lihe Yang, Bingyi Kang, Zilong Huang, Xiaogang Xu, Jiashi Feng, and Hengshuang Zhao.
\newblock Depth anything: Unleashing the power of large-scale unlabeled data.
\newblock In \emph{Proceedings of the IEEE/CVF Conference on Computer Vision and Pattern Recognition}, pages 10371--10381, 2024{\natexlab{a}}.

\bibitem[Yang et~al.(2024{\natexlab{b}})Yang, Kang, Huang, Xu, Feng, and Zhao]{yang24depth}
Lihe Yang, Bingyi Kang, Zilong Huang, Xiaogang Xu, Jiashi Feng, and Hengshuang Zhao.
\newblock Depth anything: Unleashing the power of large-scale unlabeled data.
\newblock In \emph{Proc. {CVPR}}, 2024{\natexlab{b}}.

\bibitem[Yang et~al.(2024{\natexlab{c}})Yang, Kang, Huang, Zhao, Xu, Feng, and Zhao]{yang24depthv2}
Lihe Yang, Bingyi Kang, Zilong Huang, Zhen Zhao, Xiaogang Xu, Jiashi Feng, and Hengshuang Zhao.
\newblock Depth anything {V2}.
\newblock \emph{arXiv}, 2406.09414, 2024{\natexlab{c}}.

\bibitem[Yao et~al.(2018)Yao, Luo, Li, Fang, and Quan]{yao2018mvsnet}
Yao Yao, Zixin Luo, Shiwei Li, Tian Fang, and Long Quan.
\newblock Mvsnet: Depth inference for unstructured multi-view stereo.
\newblock In \emph{Proceedings of the European conference on computer vision (ECCV)}, pages 767--783, 2018.

\bibitem[Yeshwanth et~al.(2023)Yeshwanth, Liu, Nie{\ss}ner, and Dai]{yeshwanth2023scannet++}
Chandan Yeshwanth, Yueh-Cheng Liu, Matthias Nie{\ss}ner, and Angela Dai.
\newblock Scannet++: A high-fidelity dataset of 3d indoor scenes.
\newblock In \emph{Proceedings of the IEEE/CVF International Conference on Computer Vision}, pages 12--22, 2023.

\bibitem[Zhang et~al.(2016)Zhang, Isola, and Efros]{zhang2016colorful}
Richard Zhang, Phillip Isola, and Alexei~A Efros.
\newblock Colorful image colorization.
\newblock In \emph{Computer Vision--ECCV 2016: 14th European Conference, Amsterdam, The Netherlands, October 11-14, 2016, Proceedings, Part III 14}, pages 649--666. Springer, 2016.

\bibitem[Zhang et~al.(2017)Zhang, Zhu, Isola, Geng, Lin, Yu, and Efros]{zhang2017real}
Richard Zhang, Jun-Yan Zhu, Phillip Isola, Xinyang Geng, Angela~S Lin, Tianhe Yu, and Alexei~A Efros.
\newblock Real-time user-guided image colorization with learned deep priors.
\newblock \emph{arXiv preprint arXiv:1705.02999}, 2017.

\bibitem[Zheng et~al.(2023)Zheng, Harley, Shen, Wetzstein, and Guibas]{zheng2023pointodyssey}
Yang Zheng, Adam~W Harley, Bokui Shen, Gordon Wetzstein, and Leonidas~J Guibas.
\newblock Pointodyssey: A large-scale synthetic dataset for long-term point tracking.
\newblock In \emph{Proceedings of the IEEE/CVF International Conference on Computer Vision}, pages 19855--19865, 2023.

\bibitem[Zhu et~al.(2017)Zhu, Xiong, Dai, Yuan, and Wei]{zhu2017deep}
Xizhou Zhu, Yuwen Xiong, Jifeng Dai, Lu Yuan, and Yichen Wei.
\newblock Deep feature flow for video recognition.
\newblock In \emph{Proceedings of the IEEE conference on computer vision and pattern recognition}, pages 2349--2358, 2017.

\end{thebibliography}
}

\clearpage
\appendix
\twocolumn[{
\centering
\Large
\textbf{\thetitle}\\
\vspace{0.5em}Supplementary Material \\
\vspace{1.0em}
\includegraphics[width=\textwidth]{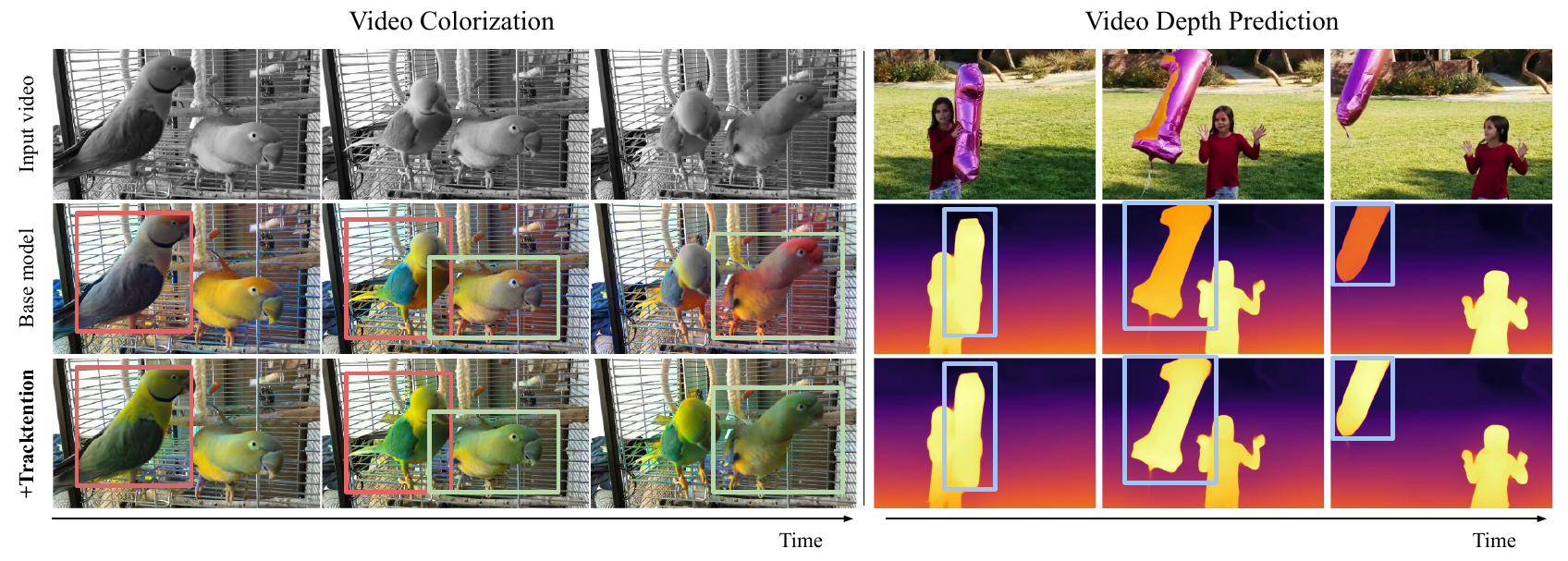}
\vspace{-2em}
\captionsetup{type=figure}
\captionof{figure}{\textbf{Stabilization effect of our model on video prediction tasks:}
\emph{Left (Video Colorization)}:
The first row shows input grayscale video frames. The second row demonstrates the output of a frame-by-frame base model~\cite{kang2023ddcolor}, producing inconsistent colors across frames (\eg, varying hues in pink and green boxes). The third row highlights our proposed Tracktention (+base model~\cite{kang2023ddcolor}), achieving consistent and stable colorization across frames.
\emph{Right (Video Depth Prediction)}:
The first row displays input video frames. The second row shows the depth predictions from a base model~\cite{yang24depth}, which suffers from temporal instability. The third row presents Tracktention's (+base model~\cite{yang24depth}) depth predictions, offering consistent and stable outputs over time.
}%
\label{fig:teaser_supp}
\vspace{1.5em}
}]

\begin{figure*}[ht]
\centering
\includegraphics[width=0.95\textwidth]{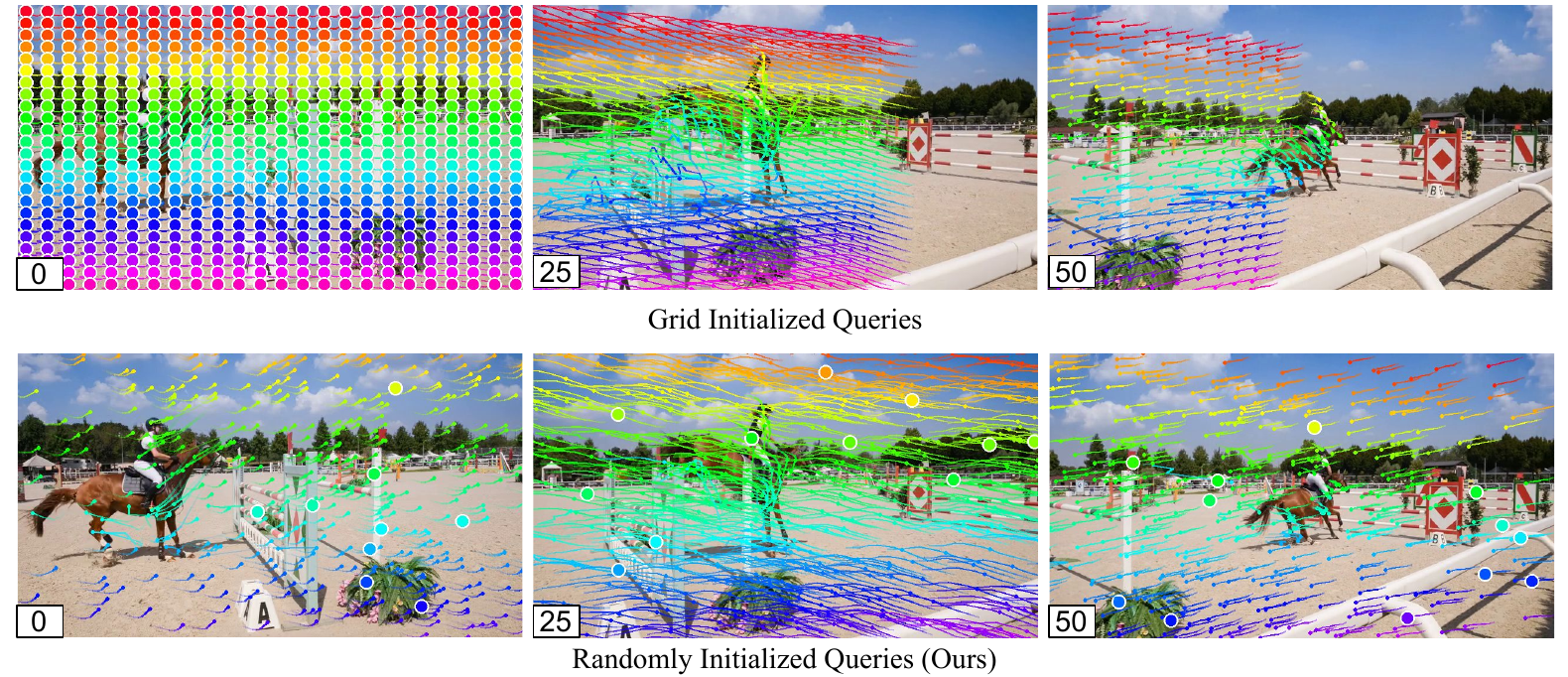}
\vspace{-1em}
\caption{\textbf{Comparison of query initialization strategies for point tracking.} The top row shows the point tracks obtained with grid-initialized queries, which suffer from significant coverage loss in later frames. In contrast, the bottom row illustrates tracks obtained with our random initialization method, which maintains comprehensive coverage across the scene over time. Larger dots with white edges represent queries, which are seed coordinates used to initiate tracking in the video. Numbers at the bottom left of each frame indicate the frame index, highlighting the improved completeness of tracks produced by our approach, particularly in later frames.}%
\label{fig:random_query}
\end{figure*}

\begin{figure}[ht]
\centering
\includegraphics[width=0.95\linewidth]{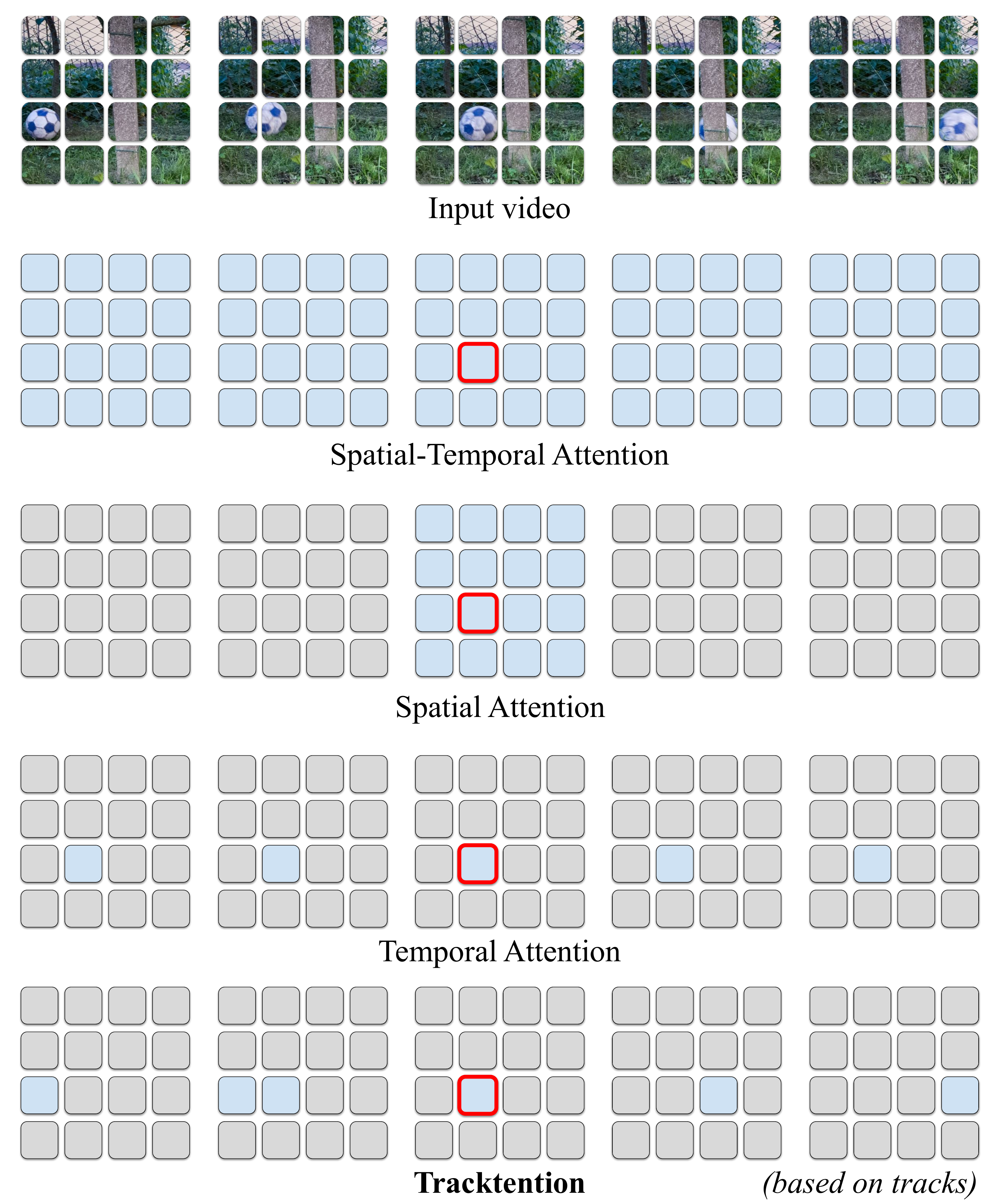}
\vspace{-1em}
\caption{\textbf{Comparison of different attention mechanisms for video processing.} The red-bordered block represents the query token, while the blue-highlighted blocks indicate the range of tokens the mechanism attends to. See text for details.
}
\vspace{-1em}
\label{fig:conceptual_comp}
\end{figure}

\section{Design and Implementation Details}

\subsection{Query Initialization and Point Tracking}

In \mname{}, we use randomly initialized queries for the point tracking model.
In Figure~\ref{fig:random_query}, we compare the results of two different query initialization strategies for object tracking in video: grid-initialized queries and random-initialized queries that we use.
Queries (represented as larger dots with white edges) are seed coordinates in the video's spatio-temporal space that are used to start the tracking process.
Effective query initialization is crucial for maintaining complete and consistent coverage of objects throughout the video frames.

The top row illustrates the tracks obtained using a grid sampling strategy, where a uniform grid of queries is placed over the spatial dimensions of the first frame.
While this method provides good initial coverage, it results in significant gaps in later frames due to motion and occlusion, leading to many areas being left untracked as the video progresses.

In contrast, the bottom row demonstrates tracks produced by our proposed random sampling method, where queries are initialized randomly in the spatio-temporal space.
This approach results in more robust and complete tracking, as seen in the wider spatial distribution of tracks in the later frames.

This enhanced tracking coverage is advantageous for our tracking-based attention model, \mname, which depends on the completeness and density of query tracks to deliver comprehensive attention across the scene.
By ensuring consistent and uniform tracking throughout the video, our random query initialization method enables \mname to more effectively focus on and process critical regions of interest, even in complex and dynamic scenarios.
This leads to improved performance and robustness, as demonstrated in the Ablation Study presented in the main paper.

\subsection{\mname compared with Standard Attention Mechanisms}

In Figure~\ref{fig:conceptual_comp}, we provide a conceptual overview comparing \mname to standard attention mechanisms for video processing.
\textbf{Spatial-temporal attention} attends to all tokens across space and time, capturing comprehensive relationships but at a prohibitive computational cost.
\textbf{Spatial attention} focuses only on spatial tokens within individual frames, ignoring temporal dependencies, while \textbf{temporal attention} processes temporal evolution at fixed spatial locations and fails when objects move across patches.
While some methods combine spatial and temporal attention, attending to a different location in another frame requires traversing intermediate tokens implicitly, leading to indirect and inefficient attention pathways.
These limitations hinder their ability to efficiently and effectively model dynamic video content.

Our proposed \mname addresses these challenges by leveraging point tracks to attend selectively to spatial-temporal tokens relevant to a query's trajectory.
Through our Attentional Sampling module, \mname handles cases where objects span multiple patches (\eg, the second frame), ensuring fragmented representations do not disrupt attention.
Additionally, it is robust to occlusion (\eg, the fourth frame) by using a tracker~\cite{karaev2024cotracker3} designed to handle occlusions.
By combining computational efficiency, adaptability to motion, and robustness to occlusion, \mname significantly improves video processing in complex scenarios.

\subsubsection{Complexity Analysis} 
We analyze the computational complexity of our proposed \mname{} layer in comparison to standard attention mechanisms used in video processing.
Let $H$ and $W$ denote the height and width of the frame features, $T$ the number of frames (temporal dimension), and $N$ the number of point tracks used in \mname{}, where $N < HW$ and $N \ll HWT$.

\paragraph{Spatio-temporal Attention:} Spatio-temporal attention attends to all tokens across both space and time, capturing comprehensive relationships but at a prohibitive computational cost: $O((HWT)^2) = O(HWT \cdot HWT)$.

This quadratic complexity over the total number of tokens $HWT$ makes it computationally infeasible for larger videos or higher resolutions.

\paragraph{Spatial Attention:} Spatial attention operates on each frame independently, attending to all spatial tokens within a frame. The computational complexity for processing all frames is: $O(T \cdot (HW)^2) = O(HWT \cdot HW)$.

This is because, for each of the $T$ frames, attention computations involve all pairs of spatial tokens, resulting in a quadratic complexity with respect to the number of spatial tokens $HW$.

\paragraph{Temporal Attention:} Temporal attention focuses on the temporal evolution at fixed spatial locations. Each spatial position attends over the temporal dimension. The computational complexity is: $O(HW \cdot T^2) = O(HWT \cdot T)$

Here, each of the $HW$ spatial positions computes attention over all $T$ temporal tokens at that position, leading to quadratic complexity with respect to the temporal length $T$.

\paragraph{\mname{} Complexity:} Our proposed \mname{} leverages point tracks to attend selectively to relevant spatial-temporal tokens along a query's trajectory. The computational complexity of \mname{} is: $O(HWT \cdot N + T^2 \cdot N)$

The term $O(HWT \cdot N)$ corresponds to the \textit{Attentional Sampling} and \textit{Attentional Splatting} process, where features are sampled along the point tracks from the video tokens.

The term $O(T^2 \cdot N)$ arises from the \textit{Temporal Transformer} operating over the point tracks, computing attention across time for each track.

\paragraph{Complexity Comparison:} Since $N$ is significantly smaller than the total number of tokens ($N \ll HWT$). The complexity of \mname{} is substantially lower than that of spatio-temporal attention mechanisms. 

Factorized spatial and temporal attention achieves better efficiency by separating spatial ($O\left(T \times (HW)^2\right)$) and temporal ($O\left(HW \times T^2\right)$) attention, resulting in a combined complexity of $O(HWT \cdot (HW + T))$. As  $N < (HW+T)$, \mname{} achieves lower complexity by attending selectively to $N$ point tracks, yielding an more efficient tool for video processing.

Furthermore, we note that the attentional sampling and splatting processes realistically focus on a local patch around each query. This allows us to further reduce the complexity of these processes from $O\left(HWT \times N\right)$ to $O\left(P^2 \times T \times N\right)$, where $P$ is a local patch size ($P^2 \ll HW$). We leave this sparsity optimization for future work.

\subsection{Other Implementation Details}

\subsubsection{Video Depth Estimation Evaluation}%
\label{app:depth_imp_details}

\paragraph{Data and evaluation metric.} For training, we use a combination of datasets containing both synthetic and real videos:
ARKitScenes~\cite{baruch2021arkitscenes},
ScanNet++~\cite{yeshwanth2023scannet++},
TartanAir~\cite{wang2020tartanair},
PointOdyssey~\cite{zheng2023pointodyssey},
DynamicReplica~\cite{karaev2023dynamicstereo},
and DL3DV~\cite{ling2024dl3dv},
totaling 12,947 videos.
All videos are resized to have a short side of 336 pixels for computational efficiency.
For evaluation, we use two benchmarks:
RobustMVD~\cite{schroppel2022benchmark}, containing short clips of 8 frames with large motion, and
the longer benchmark videos from DepthCrafter~\cite{hu2024depthcrafter}.
These evaluation datasets include a wide variety of scenes from
KITTI~\cite{geiger2013vision},
ScanNet~\cite{dai2017scannet},
DTU~\cite{yao2018mvsnet},
Tanks and Temples~\cite{knapitsch2017tanks},
ETH3D~\cite{Schops_2017_CVPR},
Sintel~\cite{mayer2016large}, and
Bonn RGB-D~\cite{palazzolo2019iros}.

We use standard depth estimation metrics~\cite{yang24depth}: \emph{Absolute Relative Difference} (AbsRel) --- calculated as $| \hat{d} - d |/d$, where $\hat{d}$ is the estimated depth and $d$ is the true depth --- and \emph{Threshold Accuracy} ($\delta_\tau$), the percentage of pixels satisfying $\max(\frac{d}{\hat{d}}, \frac{\hat{d}}{d}) < \tau$, with $\tau$ set per the original benchmark.
As in training, we calibrate each prediction to the ground truth using a scale and shift factor, shared across all frames in a video, before assessing a metric.

\paragraph{Scale and shift ambiguity}
Evaluating depth estimation in video sequences presents unique challenges due to the scale ambiguity inherent in monocular depth prediction, where predicted depths may differ from ground truth by a global scale and shift. Traditional methods~\cite{luo2020consistent, kopf2021robust, wang2023neural, shao24learning} often employ frame-wise evaluation, fitting a separate scale and shift for each frame independently:
\[
\min_{s_i,t_i} \sum_{(x, y)} \left( D_{2,i}(x, y) - \left( s_i \cdot D_{1,i}(x, y) + t_i \right) \right)^2,
\]
where 
$D_{1,i}$ is the predicted depth, 
$D_{2,i}$ is the ground-truth depth for frame $i$,
and $s_i$, $t_i$ are the scale and shift for that frame. 
While this approach minimizes per-frame error, it can mask significant temporal inconsistencies, as it allows scale and shift to vary freely between frames, leading to flickering or unstable depth predictions in videos.

To address this issue, we adopt a video-wise evaluation method similar to DepthCrafter~\cite{hu2024depthcrafter}. This approach enforces a single global scale and shift across the entire video sequence:
\[
\min_{s,t} \sum_{i} \sum_{(x, y)} \left( D_{2,i}(x, y) - \left( s \cdot D_{1,i}(x, y) + t \right) \right)^2.
\]
By using consistent scaling factors $s$ and $t$ for all frames, the video-wise evaluation penalizes variations in predicted depth over time, providing a more accurate assessment of temporal consistency. This method highlights the model's ability to maintain stable and coherent depth predictions across frames, which is crucial for applications requiring consistent video outputs.

\subsubsection{Implementation Details for Video Colorization}%
\label{app:color_imp_details}

\paragraph{Evaluation metrics. }
We assess the quality of the colorization using standard metrics:
the \emph{Fréchet Inception Distance (FID)}, which evaluates how well the predicted colorization matches the ground truth statistically in feature space;
the \emph{Colorfulness Score (CF)}, which quantifies color vibrancy; and
the \emph{Color Distribution Consistency (CDC)}, which measures the temporal consistency of the colorization.
We also include the Peak Signal-to-Noise Ratio (PSNR), though it is generally acknowledged that this is a poor metric for evaluating colorization accurately~\cite{kang2023ddcolor}.

\paragraph{Integration implementation}
We evaluate the effectiveness of the \mname{} layer in enhancing temporal consistency by integrating it into four image colorization models: CIC~\cite{zhang2016colorful}, IDC~\cite{zhang2017real}, ColorFormer~\cite{ji2022colorformer}, and DDColor~\cite{kang2023ddcolor}.

\paragraph{Integration with CIC and IDC}
For the ConvNet-based architectures CIC and IDC, which feature encoder-decoder structures with downsampling, standard, and upsampling convolutional layers, we insert the \mname{} layer after the standard convolutional layers 5, 6, and 7. This integration allows temporal alignment at multiple stages of feature extraction, enhancing consistency across video frames.

\paragraph{Integration with ColorFormer and DDColor}
In ColorFormer, which utilizes a transformer backbone followed by a 4-layer U-Net decoder with residual connections, we integrate the \mname{} layer after the first three layers of the transformer backbone. Similarly, for DDColor, which employs a ConvNeXt~\cite{liu2022convnet} backbone with a U-Net decoder, we add the \mname{} layer after the first three layers of the backbone. 

\paragraph{Training Details}
All models are trained using the loss functions from DDColor with the AdamW optimizer (initial learning rate $1.6 \times 10^{-5}$). The learning rate is decayed using a MultiStepLR scheduler every 4k iterations starting from the 8k-th iteration, over a total of 40k iterations. Training is conducted on the YouTube-VIS~\cite{Yang2019vis} dataset using raw, unlabeled video data, with frames resized to $256 \times 256$ pixels.

We do not employ temporal consistency losses such as flow warping losses used in prior works~\cite{lei2019fully, liu2021temporally}. By avoiding the computation of optical flow and associated warping operations, our training process remains efficient while achieving temporal consistency through the \mname{} layer integration. This shows that the \mname{} layer can effectively enhance temporal consistency in video colorization without relying on additional temporal loss functions, highlighting its flexibility and practicality.

\begin{figure*}[ht]
\centering
\includegraphics[width=0.95\textwidth]{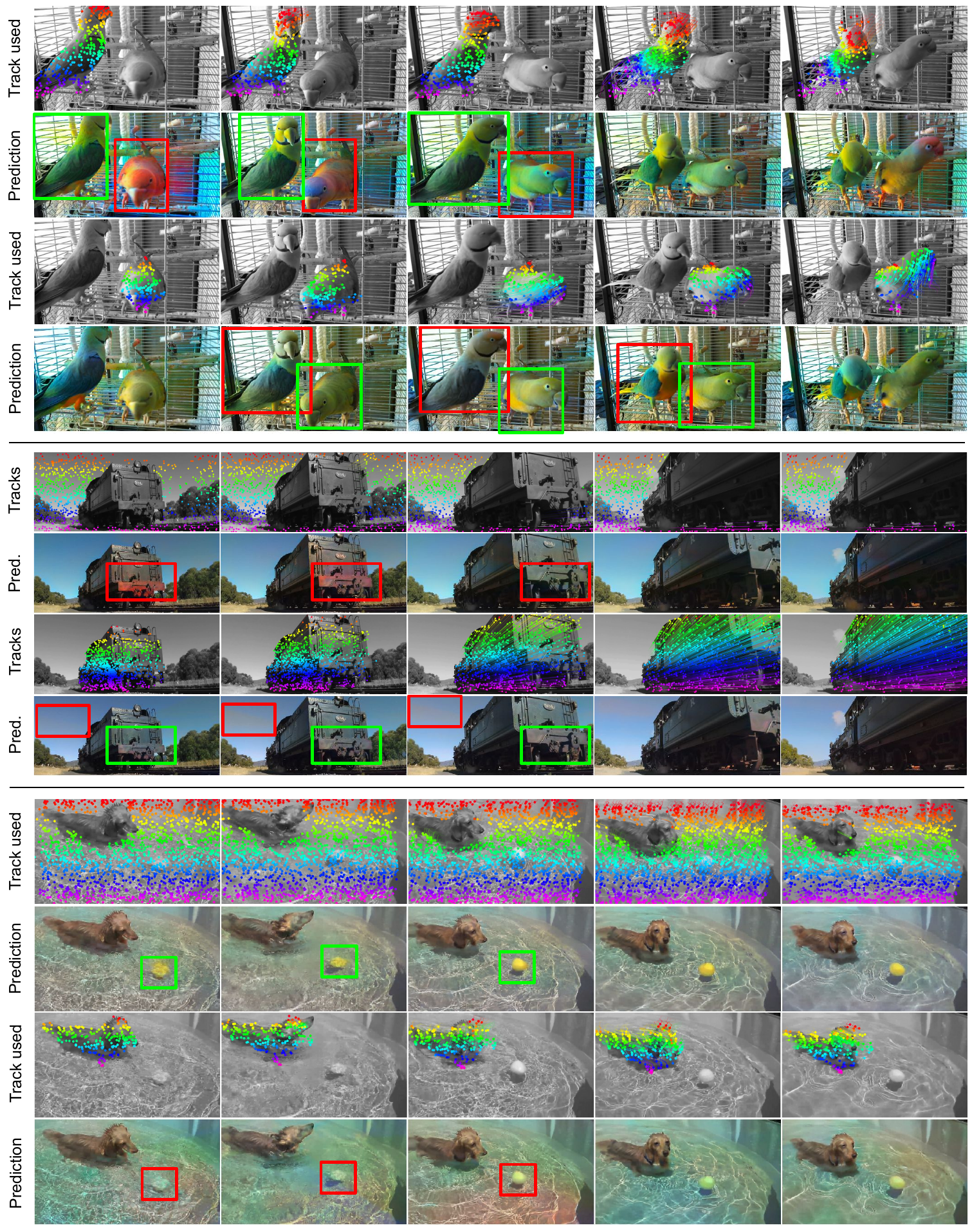}
\vspace{-1mm}
\caption{\textbf{The impact of selective tracks used on video colorization consistency.} In the top example, using tracks corresponding to the left bird (Row 1) results in consistent colorization of the left bird (Row 2, green box), while the right bird exhibits inconsistent colorization (red box). Conversely, using tracks for the right bird (Row 3) ensures stable colorization for the right bird (Row 4, green box), while the left bird becomes inconsistent (red box). Similar patterns are observed for the train and dog examples. Green boxes highlight regions with stable and consistent colors, while red boxes indicate inconsistent colorization. }%
\label{fig:track_ablate}
\end{figure*}

\begin{figure*}[ht]
\centering
\includegraphics[width=0.98\textwidth]{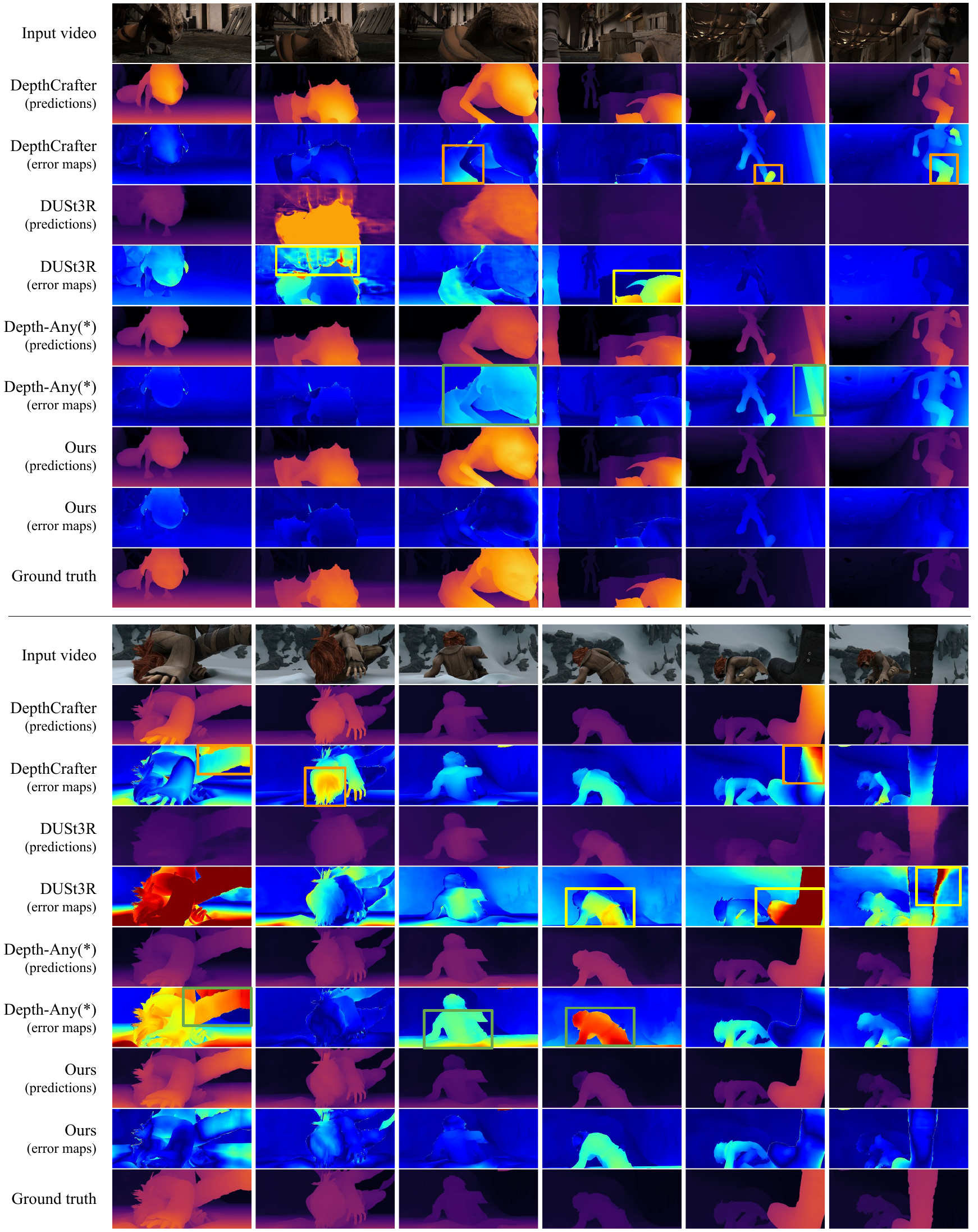}
\vspace{-0.8em}
\caption{\textbf{Comparison of depth prediction results across different models:} DepthCrafter, DUSt3R, Depth-Anything (marked with * as the base model), and Ours (+Depth-Anything). Each row shows depth predictions, corresponding error maps, and the ground truth for the input video frames. Highlighted rectangles emphasize key issues in baseline methods: DepthCrafter exhibits significant errors in certain areas, DUSt3R tends to fail in dynamic regions, and Depth-Anything produces flickering results, evident from inconsistent error patterns.}%
\label{fig:supp_fig_depth_2}
\end{figure*}

\begin{figure*}[ht]
\centering
\includegraphics[width=0.95\textwidth]{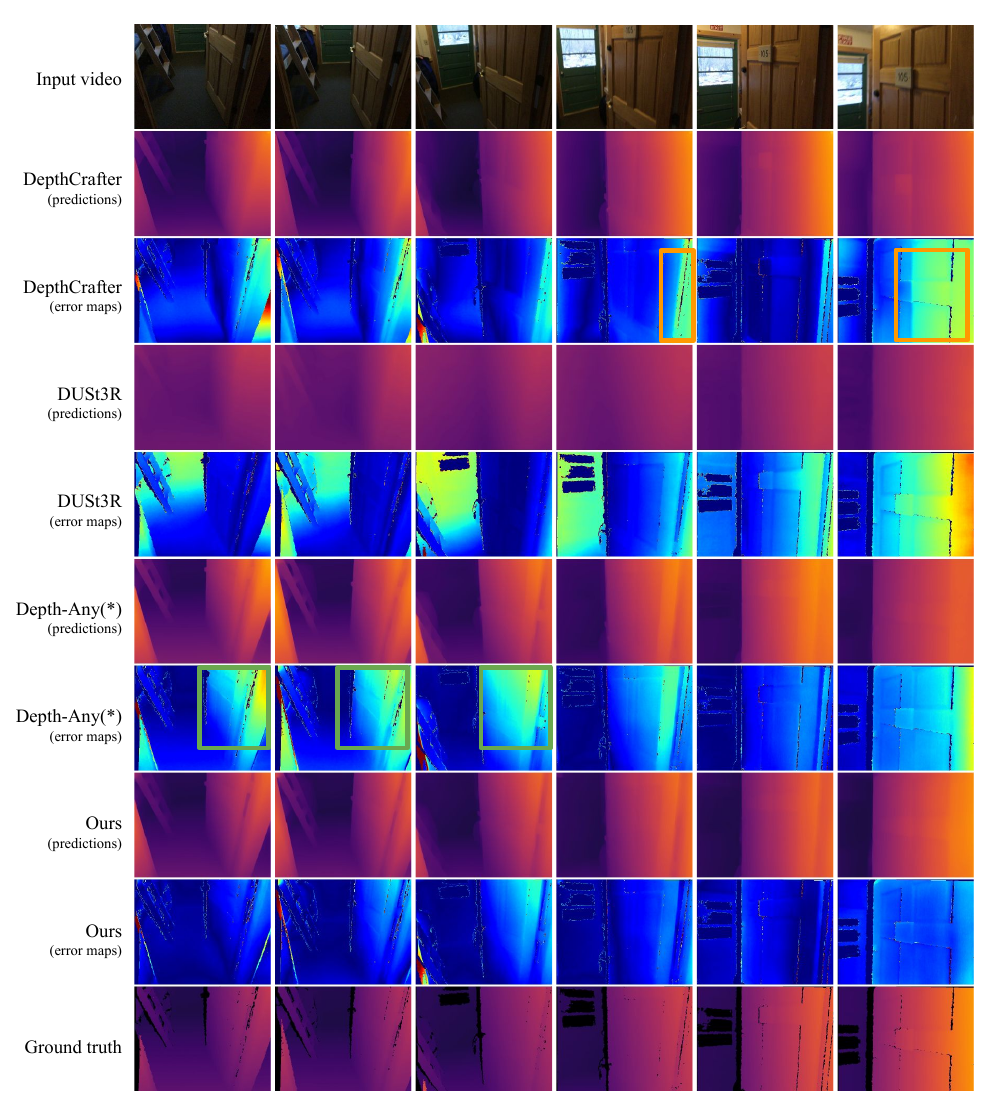}
\caption{\textbf{Additional comparison of depth prediction results} across DepthCrafter, DUSt3R, Depth-Anything (*denotes the base model), and Ours (+Depth-Anything). The rows present depth predictions, their error maps, and ground truth for a different set of input video frames, illustrating consistency across varying scenes.}%
\label{fig:supp_fig_depth_3}
\end{figure*}

\begin{figure*}[ht]
\centering
\includegraphics[width=0.95\textwidth]{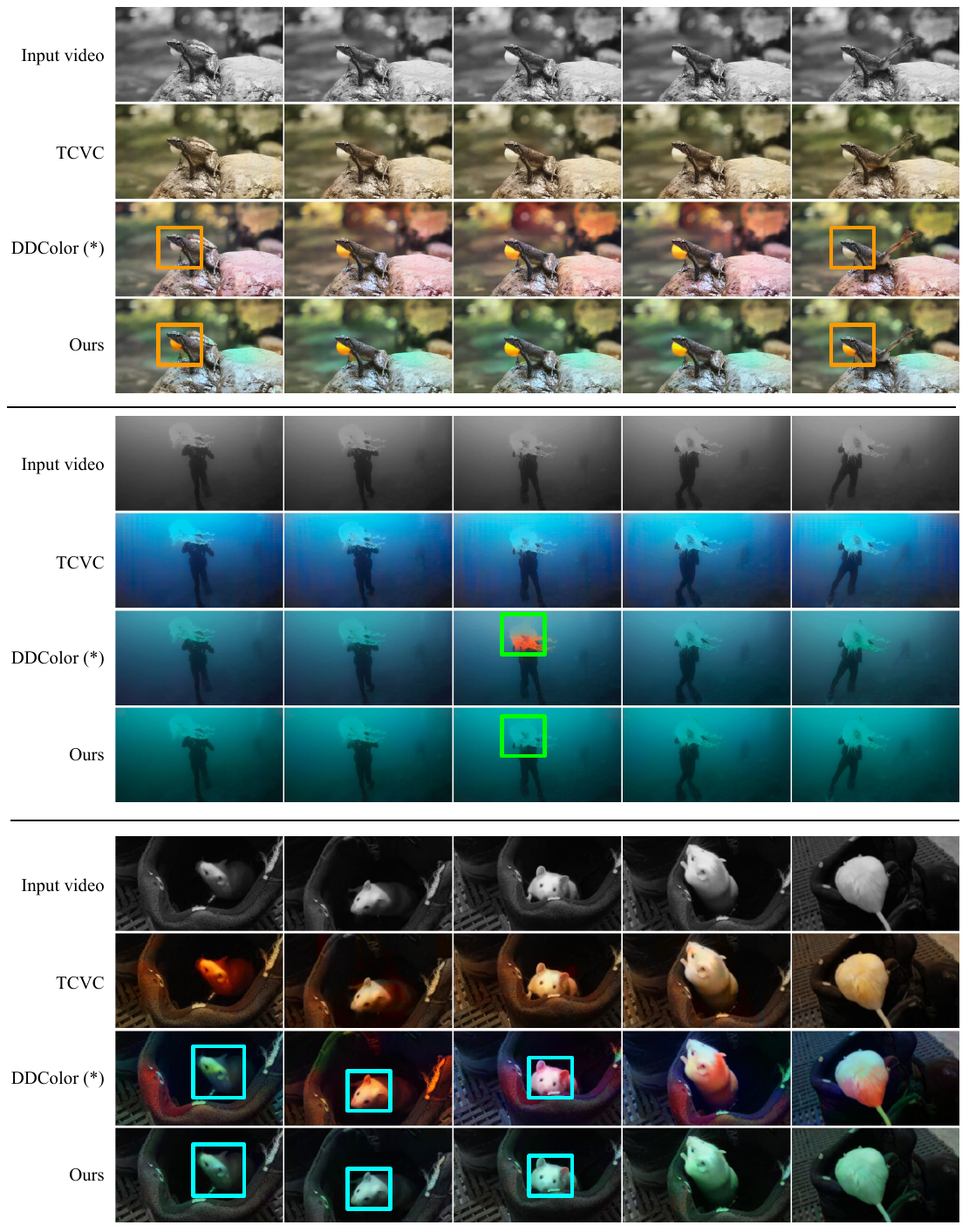}
\caption{\textbf{Video colorization results} comparing TCVC, DDColor (*denoting the base model), and Ours (+DDColor). The rows show the input grayscale video frames, followed by the colorized outputs from each method. Highlighted areas indicate inconsistencies in the base model (DDColor), which are resolved by our model, demonstrating its ability to produce consistent and accurate colorization.}
\label{fig:supp_fig_colorization_1}
\end{figure*}

\begin{figure*}[ht]
\centering
\includegraphics[width=0.95\textwidth]{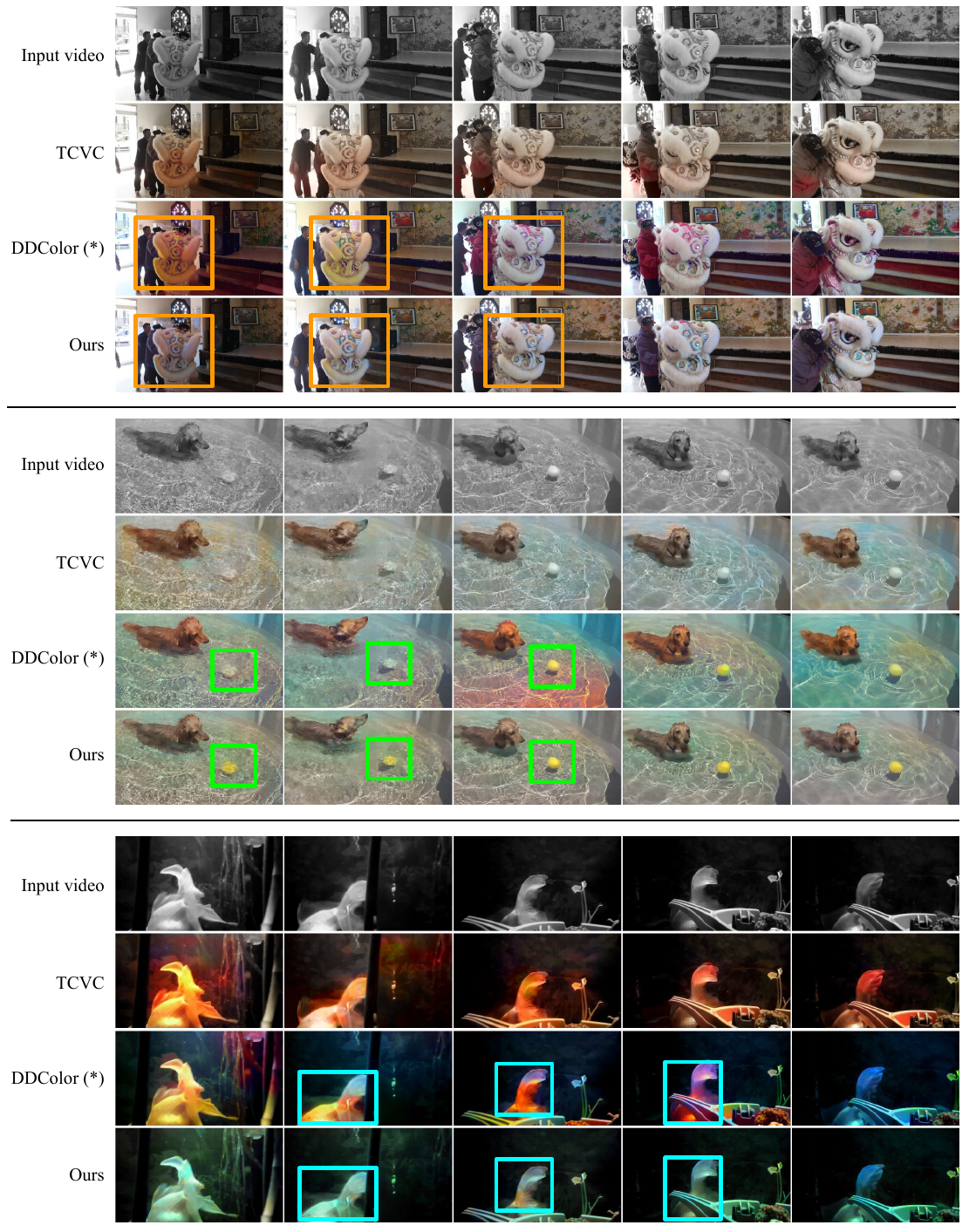}
\caption{\textbf{Additional video colorization results} comparing TCVC, DDColor (*denoting the base model), and Ours (+DDColor). The rows display the input grayscale video frames alongside the colorized outputs from each method. Highlighted areas pinpoint inconsistencies in the base model (DDColor), which are effectively resolved by our model, showcasing its improved consistency and color accuracy.}%
\label{fig:supp_fig_colorization_2}
\end{figure*}

\section{Additional Experimental Results}
Here, we present additional experimental results and visual comparisons showing how the \mname{} layer enhances video consistency, stability, and accuracy, along with an analysis of input tracks' influence.

\subsection{Influence of Input Tracks on Video Consistency}
We analyzed how the selective use of specific tracks affects the temporal consistency of video colorization. By selectively activating tracks corresponding to specific objects or regions, \mname{} guides its attention mechanism to maintain stable and consistent colorization for those targeted areas across video frames, while areas without active tracks may exhibit color inconsistencies. Figure~\ref{fig:track_ablate} illustrates this behavior. In each case, activating tracks for a particular object (e.g., one bird, the front of the train, or the dog) results in consistent coloring for that object, while other objects without active tracks (e.g., the other bird, the back of the train, or the ball) show inconsistent coloring. This demonstrates that \mname{} can leverage selective tracks to ensure localized stability in the colorization process, even when other regions of the frame are affected.

\subsection{Additional Visual Examples}
\paragraph{Video Depth Estimation}
Figures~\ref{fig:supp_fig_depth_2} and~\ref{fig:supp_fig_depth_3} present additional visual comparisons for video depth estimation. Similar to the results in the main paper, our model produces stable and accurate depth maps across all frames. The incorporation of the \mname{} layer enables precise temporal alignment of features, resulting in consistent and accurate depth estimation over time.

The state-of-the-art method, DepthCrafter exhibits significant errors in certain regions, particularly where complex motion or occlusions occur. DUSt3R, which relies on implicit triangulation, struggles with dynamic content, leading to inaccuracies and temporal inconsistencies in the estimated depth maps.

Our base model, Depth Anything, is an image-based depth estimation model that processes frames independently. As a result, it shows inconsistent depth estimation across frames, with noticeable instability in the depth maps. By integrating the \mname{} layer into Depth Anything, we enhance temporal consistency, achieving results that are both accurate and stable throughout the video sequence.

\paragraph{Automatic Video Colorization}
Figures~\ref{fig:supp_fig_colorization_1} and~\ref{fig:supp_fig_colorization_2} provide more results on automatic video colorization. Consistent with observations in the main paper, the baseline method TCVC is unable to produce vibrant colors, resulting in desaturated and less realistic outputs. 

Our base model, DDColor, performs frame-by-frame colorization, which leads to unstable and inconsistent color results. The absence of temporal coherence causes color flickering and discrepancies between frames, detracting from the overall visual quality of the video.

When we augment DDColor with our \mname{} layer, we observe a clear improvement in temporal consistency while retaining the original color vibrancy and realism. The \mname{} layer allows the model to attend to corresponding areas across time based on point tracks, ensuring smooth and coherent colorization throughout the video.

\subsection{Video Demonstrations}
We provide a supplementary video that compares base models with and without our \mname{} layer for video depth estimation and colorization, highlighting improved temporal consistency and accuracy. See \url{youtu.be/xuEM_Y4MO6E} file for the video.

\end{document}